\documentclass[final,1p,times]{elsarticle}
\usepackage{verbatim} 
\usepackage{graphicx}
\usepackage{amssymb}
\usepackage{amsthm}
\usepackage{amsmath}
\usepackage{nccmath}
\usepackage{amsbsy}
\usepackage{lineno}
\usepackage[nolist]{acronym}
\usepackage{amsfonts}
\usepackage{multirow}
\usepackage[table]{xcolor}
\journal{Journal Name}
\usepackage{mathtools}
\usepackage{svg}
\usepackage{tablefootnote}

\usepackage[ruled,vlined,linesnumbered]{algorithm2e}
\usepackage{todonotes}
\theoremstyle{definition}

\usepackage{lscape}
\usepackage{placeins}

\usepackage{nomencl}
\usepackage[flushleft]{threeparttable}
\usepackage{url}
\usepackage{resizegather}
\usepackage{eufrak}
\usepackage{mathtools}
\usepackage{lscape}
\usepackage{epstopdf}

\journal{Journal}

\biboptions{authoryear}

\begin{document}
\begin{frontmatter}

\title{Robust Unsupervised Domain Adaptation by Retaining Confident Entropy via Edge Concatenation}

\author[label1]{Hye-Seong Hong}
\ead{ghdtjd0810@knu.ac.kr}

\author[label1]{Abhishek Kumar}
\ead{abhishek.ai@knu.ac.kr}

\author[label1]{Dong-Gyu Lee \corref{cor1}}
\ead{dglee@knu.ac.kr}

\cortext[cor1]{Corresponding author.}
\address[label1]{Department of Artificial Intelligence, Kyungpook National University, Daegu, South Korea}

\begin{abstract}
The generalization capability of unsupervised domain adaptation can mitigate the need for extensive pixel-level annotations to train semantic segmentation networks by training models on synthetic data as a source with computer-generated annotations. 
Entropy-based adversarial networks are proposed to improve source domain prediction; however, they disregard significant external information, such as edges, which have the potential to identify and distinguish various objects within an image accurately. 
To address this issue, we introduce a novel approach to domain adaptation, leveraging the synergy of internal and external information within entropy-based adversarial networks. In this approach, we enrich the discriminator network with edge-predicted probability values within this innovative framework to enhance the clarity of class boundaries. Furthermore, we devised a probability-sharing network that integrates diverse information for more effective segmentation. Incorporating object edges addresses a pivotal aspect of unsupervised domain adaptation that has frequently been neglected in the past-- the precise delineation of object boundaries. Conventional unsupervised domain adaptation methods usually center around aligning feature distributions and may not explicitly model object boundaries. Our approach effectively bridges this gap by offering clear guidance on object boundaries, thereby elevating the quality of domain adaptation. Our approach undergoes rigorous evaluation on the established unsupervised domain adaptation benchmarks, specifically in adapting SYNTHIA $\rightarrow$  Cityscapes and SYNTHIA $\rightarrow$  Mapillary. Experimental results show that the proposed model attains better performance than state-of-the-art methods. The superior performance across different unsupervised domain adaptation scenarios highlights the versatility and robustness of the proposed method. 
\end{abstract}


\end{frontmatter}
\section{Introduction}\label{introduction}
Semantic segmentation plays a crucial role in the field of computer vision, enabling a comprehensive understanding of images by analyzing both their content and context, contributing to high-level image understanding. 
The primary objectives of semantic segmentation include labeling image pixels with classes and defining object boundaries~\citep{hoffman2016fcns}. This helps in comprehending relationships and the visual appearance of objects. Its applications span diverse domains, including depth estimation~\citep{bhat2021adabins}, object detection~\citep{zhao2019object}, and image segmentation~\citep{minaee2021image}, ~\citep{kumar2022semantic}, providing rich outputs. The importance of semantic segmentation extends to medical imaging~\citep{asgari2021deep}, autonomous driving~\citep{lee2022joint}, ~\citep{lee2021fast}, aerial crop monitoring~\citep{anand2021agrisegnet}, and point cloud segmentation~\citep{guo2020deep}. Deep learning advancements enhance semantic segmentation, but it relies on pixel-wise extensive annotated data. Manual pixel-level annotation is labor-intensive, and even semi-automatic tools require validation of generated labels; therefore, acquiring labeled data poses challenges. Ensuring high performance in new domains is elusive when networks train on domain-specific datasets, as disparities in data distributions, like weather conditions, affect performance. Gathering diverse data and training can help but demands significant effort. 

Recent advancements in computer graphics and gaming technologies have opened up the opportunity to construct lifelike virtual environments encompassing a wide range of landscapes, convincing characters, and authentic behaviors. Within the machine learning and computer vision applications, these tools have been recognized for their utility in generating datasets to train deep learning networks~\citep{richter2016playing}. The ever-increasing realism of these artificially generated datasets allows for creating an extensive pool of labeled data, enhancing the efficacy of models when applied to real-world data. Moreover, this approach is pivotal in addressing the challenge of rare and extreme real-world situations. Collecting data for extremely rare events, such as accidents involving a hundred cars or other uncommon occurrences, is not only logistically challenging but also ethically and financially impractical. In such cases, leveraging the versatility of computer-generated environments provides a unique advantage. By simulating these rare and extreme scenarios within the controlled virtual space, researchers can generate datasets with detailed labels for small object that are essential for training models to handle such exceptional situations. Additionally, this approach facilitates the creation of diverse datasets by simulating various scenarios; for instance, it becomes possible to simulate changing weather conditions and architectural styles when simulating driving scenarios. This versatile process provides an adaptable platform for developing and training computer vision models for various tasks, including semantic segmentation (e.g., ~\citep{zhang2017curriculum}).

In practice, simulating images can train semantic segmentation models to maintain some level of performance on real images~\citep{zhang2021survey}. However, domain adaptation can enhance this performance further significantly. Unlike supervised training, domain adaptation doesn't rely on labeled data for the target domain, making it a valuable approach~\citep{ganin2015unsupervised}. At its core, domain adaptation leverages labeled data from relevant source domains to boost the model's performance in the target domain, employing the versatile concept of Transfer Learning~\citep{long2015learning}. Domain adaptation methods play a pivotal role in addressing domain shifts and distribution changes, which are known to degrade model performance, as extensively discussed in the literature~\citep{jiang2008literature}. Practitioners can integrate target domain samples into the training process to facilitate domain adaptation or rely solely on source domain data. Notably, the source-only approach, which relies exclusively on annotated source domain data, often falls short in the face of even minor domain shifts. In contrast, Unsupervised Domain Adaptation (UDA) models take a more comprehensive approach by leveraging both labeled source and unlabeled target domain samples (known as source-to-target UDA). This approach involves a form of supervised learning within a virtual environment, followed by adaptation to the target domain conditions within that same environment, resulting in improved model generalization.

Adversarial learning-based methods have shown their effectiveness in aligning features for semantic segmentation tasks. This has been highlighted in several studies~\citep{murez2018image, hoffman2018cycada, tsai2019domain, pan2020unsupervised}, which focus on aligning features at either the image level, the output level, or both. Most of these earlier works introduced adversarial networks in the context of generative models~\citep{goodfellow2020generative,saporta2020esl}. In a more advanced direction, there has been an exploration into using entropy probability~\citep{vu2019advent} to align different domain distributions. Depth context information has also been leveraged to enhance performance~\citep{vu2019dada, saha2021learning}. In \cite{vu2019dada}, depth information was decoded into pixel-wise values, enabling its integration with semantic information. The study by \cite{saha2021learning}, known as CTRL, introduced an entropy module using discretized depth distributions, deviating from the direct combination of semantic and depth information features.

Nonetheless, prior studies~\citep{murez2018image, hoffman2018cycada, tsai2019domain, pan2020unsupervised} often overlooked the holistic object outline information, focusing primarily on semantic region details derived from depth and semantics. In human perception of the object, context alone, such as depth, brightness, or color, doesn't suffice. Human perception hinges on capturing object outlines and subsequently inferring the overall RGB characteristics, akin to how images evolve from initial outlines to complete RGB representations. Essentially, edge and semantic information coalesce in the intricate process of visual perception. Recent strides owe their debt to earlier research in UDA, which harnesses entropy-driven techniques, combined with recent contributions like the CTRL~\citep{saha2021learning} study. This amalgamation has yielded outstanding UDA performance through entropy minimization in semantic probability values. When confronted with intricate object boundaries, this approach generates heightened entropy values. It's crucial to emphasize that these values signify persistent predictive uncertainty encircling object boundaries rather than authentic edges. In response to this insightful observation, we propose a pioneering approach by introducing object edges into the framework of UDA. This represents a novel contribution to the existing literature, as, to the best of our knowledge, the incorporation of object edges within UDA has not been explored previously. We envision that this innovative inclusion of object edges offers a dual advantage for UDA applications. First and foremost, integrating object edges into the UDA process can significantly enhance the overall effectiveness of entropy-based techniques. By utilizing object edges as unequivocal markers of object boundaries, we introduce a new dimension to the entropy-driven adaptation process. These edges serve as crucial cues, guiding the model in accurately delineating objects and facilitating a more precise alignment between the source and target domains. Furthermore, the introduction of object edges addresses a critical aspect of UDA that has often been overlooked in the past—the delineation of object boundaries. Traditional UDA methods typically concentrate on feature distribution alignment but may lack the ability to model object boundaries explicitly. Our approach bridges this gap by providing definitive guidance on object boundaries. This improves the quality of domain adaptation and opens up new possibilities for downstream tasks, such as object segmentation and recognition.

    

\begin{figure}[t]
\centering
\includegraphics[width=0.9\linewidth]{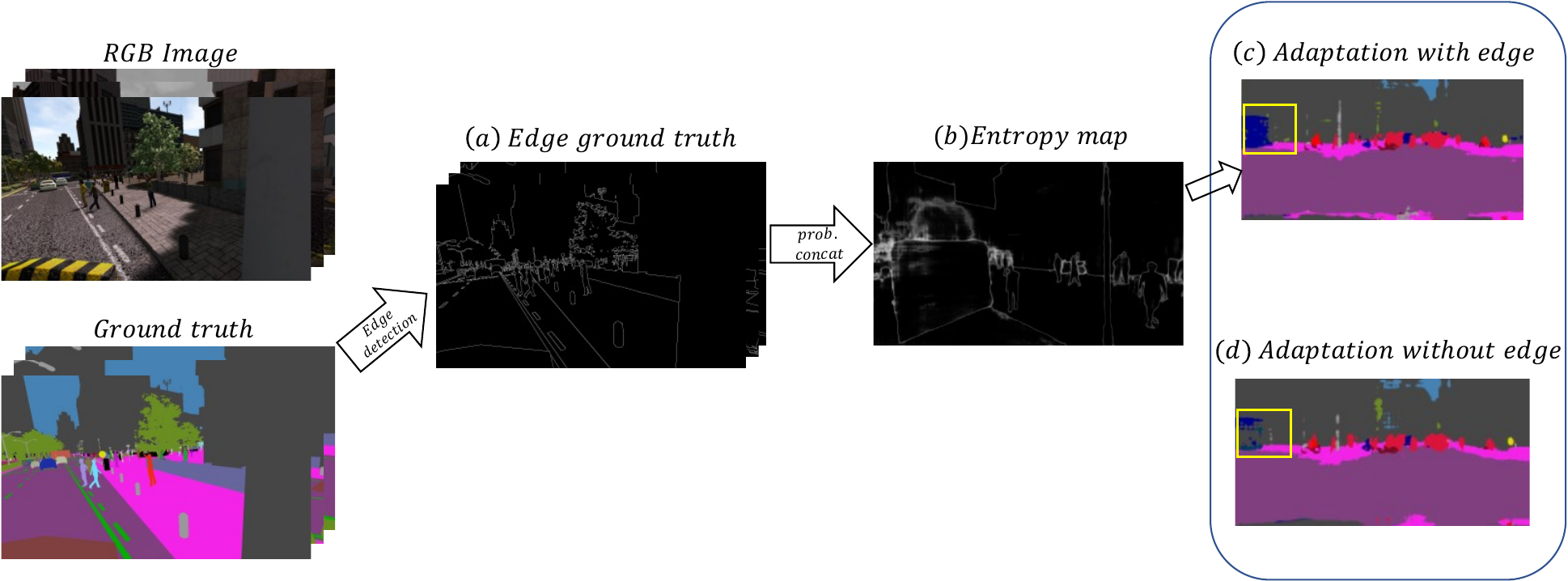}
\caption{After extracting edge label from ground truth, concatenate predicted probability to depth and semantic entropy. (a) Show edge ground truth. (b) Refers to an entropy map. (c) and (d) Show different visualizations of adaptation results.}
\label{1_intro}
\end{figure}

In this paper, we leverage both edge and semantic information within an adversarial network framework. We incorporate predicted edge probabilities into the discriminator network. Including edge information is a valuable strategy for enhancing the clarity of class boundaries. For instance, consider a scenario where the probability distributions for three classes at a single pixel are $0.4$, $0.3$, and $0.3$. Employing predicted entropy minimization for internal information may adjust these probabilities to $0.6$, $0.2$, and $0.2$. However, introducing supplementary edge information to the discriminator yields the advantage of a more distinct probability distribution. The presence of ground truth edge data, along with the successful outcomes achieved by incorporating concatenated edge information, is demonstrated in Figure \ref{1_intro} of this paper. We employ Canny edge detection on the semantic segmentation ground truth map to establish edge labels, resulting in a comprehensive edge ground truth map for each label class. Additionally, we devise a probability-sharing network that considers different types of information for the segmentation task. Further elaboration on this process will be provided in the forthcoming methodology section. To summarize, the key contributions of our work are as follows: \color{black}

\unskip

\begin{itemize}
  \item  \textit{Innovative Dataset Generation}: We introduce a novel approach to dataset creation by extracting class-specific edge labels directly from segmentation maps, fostering a more efficient and tailored data resource.
  \item \textit{Enhanced Information Transmission}: Our methodology achieves concurrent transmission of edge probability values alongside depth and semantic probability values to the discriminator during domain adaptation training, enhancing the overall model's capacity to assimilate and adapt to diverse data domains. 
  \item \textit{Performance Enhancement}: Our last key contribution lies in achieving significantly improved evaluation accuracy compared to the previously established method when applied to the synthetic-2-real benchmark datasets.
\end{itemize}
The remaining sections are organized as follows: Section 2 delves into the relevant literature. Section 3 provides a concise overview of the proposed design. Section 4 presents and analyzes the results of the experiments conducted. Finally, in Section 5,  we conclude and wrap up the discussion.
\section{Related Work}

\subsection{Semantic Segmentation}
Semantic segmentation is a computer vision problem that involves assigning each pixel in an image to a specific class. Usually, this issue is tackled using supervised training, which involves training a machine-learning model on an image dataset labeled at the pixel level. Semantic image segmentation aims to identify the semantic meaning of distinct segments of an image. It should be noted that low-level image segmentation, which is the unsupervised partitioning of an image into coherent regions based on primary indications, including texture, depth, or color, is a relevant but distinct problem not addressed in this survey.

Semantic segmentation models based on deep learning are extensive and can be categorized into several groups based on their primary principles \citep{minaee2021image}. The following section provides a brief overview of the most common approaches. However, due to the vast literature on semantic segmentation, we recommend that interested readers refer to the \cite{minaee2021image}, a survey paper, for a more comprehensive review.
Recently, convolutional neural networks have grown rapidly due to the development of deep learning, and AlexNet \citep{krizhevsky2017imagenet}, VGG \citep{simonyan2014very}, and ResNet \citep{he2016deep} are representative models of convolutional neural networks. Semantic segmentation is one of the application fields, which is the task of assigning semantic labels to each image pixel. UNet \citep{ronneberger2015u} performed a semantic segmentation task with a structure designed with an encoder-decoder. After fully convolutional networks in semantic segmentation \citep{long2015fully} are first introduced, numerous semantic segmentation models are developed \citep{chen2017deeplab, zhao2017pyramid, yu2015multi}.

Another category of semantic segmentation methods involves Recurrent Neural Networks (RNNs) instead of Convolutional Neural Networks. Recurrent neural networks are particularly effective at modeling long-distance dependencies among pixels, which can improve segmentation quality. For example, Visin \textit{et al.} proposed the ReNet architecture, which consists of four RNNs per layer and is effective in semantic segmentation \citep{visin2016reseg}. Similarly, Byeon et al. introduced two-dimensional LSTM networks that capture complex spatial dependencies among labels \citep{byeon2015scene}. Another approach is the Graph LSTM model proposed by Liang \textit{et al.}, which treats arbitrary-shaped superpixels as nodes in a graph and their spatial relationships as edges, resulting in a more semantically consistent segmentation \citep{liang2016semantic}.

Several recent works explored the use of transformer models for semantic segmentation. In \cite{strudel2021segmenter}, an enhanced variant of Vision Transformer (ViT) models \citep{dosovitskiy2020image} is proposed to deal with semantic segmentation tasks. Unlike CNN-based models, ViT-based models can capture global context information from the very beginning, which helps in improving segmentation quality. Strudel \textit{et al.} rely on output embeddings of image patches and use either a mask transformer decoder or a point-wise linear decoder to obtain class labels~\citep{strudel2021segmenter}. In \cite{xie2021segformer},  hierarchical transformer-based encoders are employed to capture overall and detailed attributes and use efficient multilayer perceptron (MLP) decoders for aggregating data from different layers. This approach combines global and local attention mechanisms to create a better robust representation. Similarly, in \cite{guo2021sotr}, a hierarchical approach is employed, which utilizes a Feature Pyramid Network to generate multi-scale feature maps and feed them into a transformer to estimate particular classes and obtain global interdependencies. They also use a multi-level upsampling unit to create segmentation maps directed by the transformer outcome. ~{\cite{ranftl2021vision}} propose a transformer model for dense forecasting,  encompassing semantic segmentation and depth evaluation.  Their approach involves utilizing the area-based result of convolutional neural networks with supplemented position embedding, constructing image-like representations at different scales by aggregating symbols from diverse phases of the vision transformer, and gradually combining them into high-resolution projections through a convolutional decoder.

\subsection{Semantic Edge Detection}
Semantic edge detection is the task of assigning an edge label to each semantic region. Edge extraction without a label may result in extracting unnecessary noise edges due to its various textures. Previous work \citep{singh2008edge} set entropy threshold in extracting grayscale edge to escape unnecessary noise edge. However, it was difficult to extract edges by semantics accurately. Other prior works \citep{yu2018simultaneous, yu2017casenet} extracted edge based on semantic edge label. Dexidet \citep{poma2020dense} tried to extract each semantic edge by grayscale edge label. However, These trials require expensive costs to build data because each semantic needs to be labeled. In this work, we introduce an idea to perform edge-supervised learning by extracting edges from semantic segmentation labels without any cost.

In the domain of computer vision, various approaches have been proposed to detect semantic edges using different techniques. One such approach was proposed by \cite{hariharan2011semantic}, which integrated standard object detectors with bottom-up contours.  \cite{bertasius2015deepedge} introduced HFL, a method that uses deep semantic segmentation networks to produce class-agnostic binary boundaries and assign class attributes to boundary edges.  \cite{maninis2016deep} integrated their Convolutional-Oriented Boundaries with semantic segmentation obtained through widened convolutions ~\citep{yu2015multi} to acquire semantic edges. In addition,  \cite{khoreva2016weakly} proposed a weakly supervised learning strategy that obtained high-quality object boundaries using only bounding box annotations without requiring any object-specific annotations. Finally,  \cite{takikawa2019gated} introduced Gated-SCNN, which significantly improves semantic segmentation by converting the semantic edge features from distinct ResNet layers to a feature appropriate in segmentation.

\subsection{Unsupervised Domain Adaptation}
UDA is a method to synchronize the distribution change among unlabeled target data and labeled source data. Lately, UDA approaches based on adversarial learning have revealed remarkable success in training domain-consistent attributes, including complex assignments such as semantic segmentation. Several studies have explored adversarial-based UDA models for semantic segmentation, including \citep{vu2019advent, chen2019domain, tsai2019domain, pan2020unsupervised, saito2018maximum, hoffman2018cycada, park2019preserving}. Typically, adversarial-based UDA models for semantic segmentation consist of two networks. The initial network acts as a generator that anticipates segmentation maps for input images from either the target or source domains. The latter network acts as a discriminator to estimate the domain labels based on features from the generator. The generator's objective is to mislead the discriminator, thereby minimizing the distributional discrepancy between the features of the two domains. In addition to feature-level alignment, some approaches attempt to synchronize the domain's variation at the image or output level. Regarding image level, CycleGAN~\citep{zhu2017unpaired} was used in \cite{hoffman2018cycada} to generate images for domain alignment.  At the output level,   \cite{pan2020unsupervised} proposed a network based on an end-to-end approach that involves structural output synchronization for distribution displacement. Most recently,  \cite{vu2019advent} utilized the pixel-wise prediction entropy from segmentation outcomes to tackle the domain disparity. \cite{stan2021unsupervised} proposed prototypical Gaussian distribution that encodes intermediate embedding space, which is aligned with target domain distribution in the embedding space. \cite{chen2021enhanced} introduced classification-constrained discriminators to enhance the traditional adversarial learning schemes to alleviate feature distortion. \cite{jiang2022prototypical} integrates inter-class information into prototypes based on classes and employs class-centered distribution alignment for the adaptation process. \cite{gao2022cross} extract the content and style knowledge contained in features and calculate the degree of inherent or illumination difference between two images to cope with domain adaptation from daytime to nighttime. Recently, the Adaptive Refining-Aggregation-Separation \citep{cao2023adaptive} framework was introduced, aiming to learn discriminative features by devising distinct adaptive strategies tailored to various domains and features. 

Since the generative adversarial network was first introduced into the world \citep{goodfellow2020generative}, Hoffman brought discriminator \citep{hoffman2016fcns} to domain adaptation fields. After that, numerous adversarial-based networks \citep{hung2018adversarial, chen2017no, du2019ssf, tsai2018learning, dong2020cscl, hoffman2018cycada} are proposed to learn the difference of domain shift. Other methods are the curriculum learning model \citep{zhang2019curriculum, pan2020unsupervised, lian2019constructing}, which are the approaches of systematical education learning where the model can fully ingest features from easy to complex. The knowledge distillation methods \citep{cai2021dlnet, hinton2015distilling} are one of the effective ways to bridge the domain gap in unsupervised learning, as the trained model transfers knowledge to the naive model in a similar way to teaching students. As geometric information is one of the essential semantic features, many researchers have conducted research to estimate depth for understanding semantic scene accurately~ \citep{vu2019dada, saha2021learning, chen2019learning, lee2018spigan}.

\begin{figure}[t]
\centering
\includegraphics[width=0.9\linewidth]{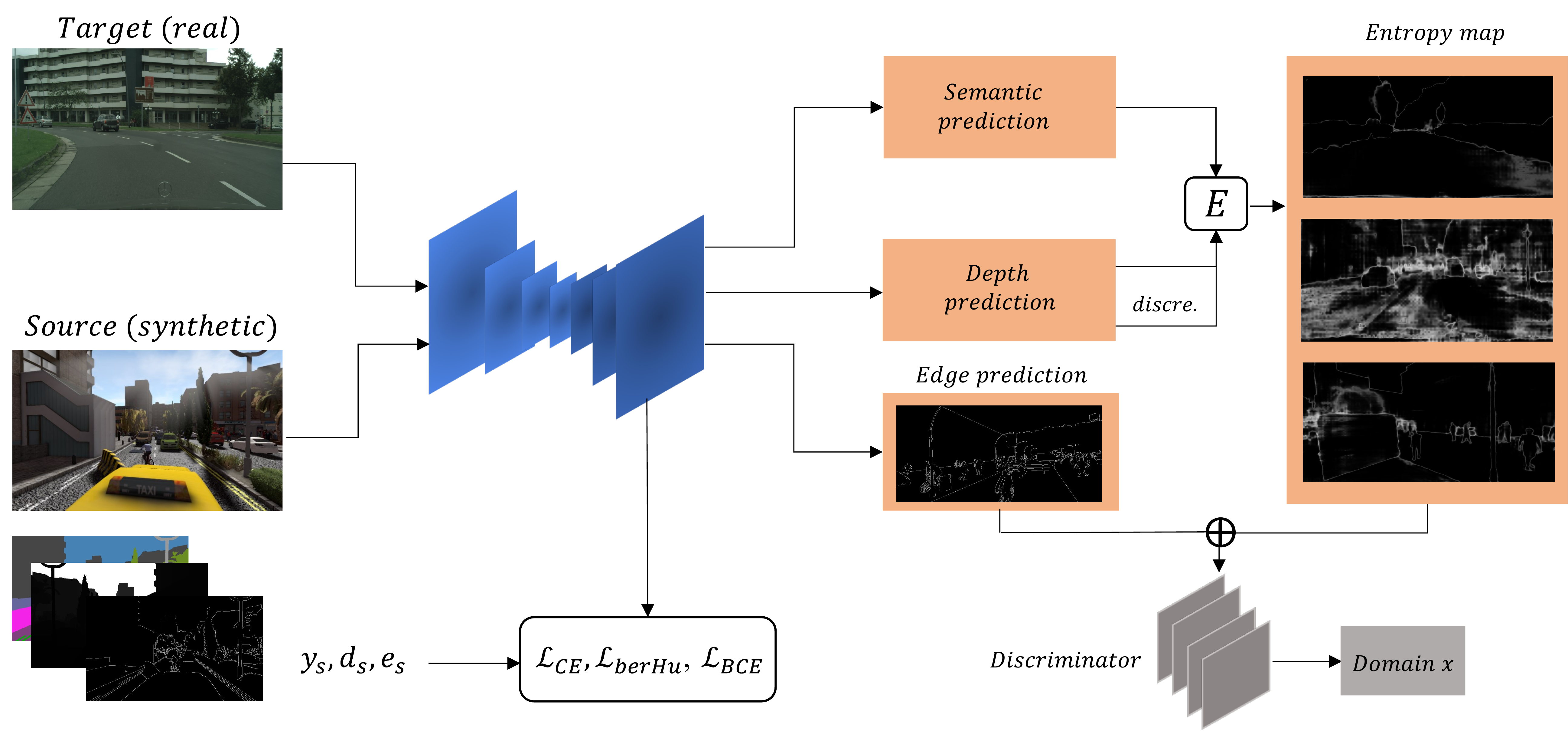}
\caption{Overview of proposed method.}
\label{architecture}
\end{figure}
\unskip

\section{Proposed Method}
\subsection{Overview}
Using existing semantic segmentation frameworks, we add a network component used for domain adaptation to build our model. Supervised learning is conducted based on DeepLabv2 with a ResNet-101 backbone and a decoder. Having four convolutional decoder layers, the output features of the decoder are obtained from the feature map of the backbone. When discrete and continuous depth prediction is combined with the CTRL scheme\citep{saha2021learning}, we can get three types of output features: semantic, depth, and edge. Depth and semantic information are fed into an entropy formulation to obtain an entropy map. Due to its uncertainty, a strong connection exists between entropy measurement and unsupervised domain adaptation. We then transfer entropy and edge prediction values from source samples to target samples to a discriminator for unsupervised domain adaptation tasks. Iterative Self Learning (ISL) training is an effective method of improving additional semantic segmentation performance. We use ISL training to exploit high-confidence predictions as supervision and our complementary edge network to predict high-quality predictions. In contrast to \cite{lee2018spigan, saha2021learning, vu2019dada} tasks that consider depth features as semantics, in this work, the emphasis is not simply on extending semantic information such as depth, but on designing the model to interact between contour and internal context information. Thus, each pixel's class probability is strengthened by the model.
To predict edge probability, we manually collect edge labels. In Figure 1, we can construct edge labels by applying Canny edge detection to ground truth for semantic classes.

\subsection{Proposed Approach}

This section illustrates our presented approach utilizing depth information and edge detection. Our base hypothesis in the UDA setting is the adversarial network for alleviating the domain gap. ${D}_S$ and ${D}_T$ are the domains of the source domain and the target domain. To organize a set of datasets, $\mathcal{D}_S\ =\ \left\{\left(x_1^S,y_1^S,z_1^S,e_1^S\right),\ \ldots,\ \ldots,\left(x_n^S,y_n^S,z_n^S,e_n^S\right)\right\},\ \ \mathcal{D}_T=\left(x_1^t,\ldots\ ,\ x_1^n\right).$ Given $X\ \in\mathbb{R}^{H\ \times\ W\ \times\ 3}$, corresponding C-class pixel annotation label follow $y\in\ {{1,....,C}}^{H\ \times\ W}$. We consider each depth and edge label $z\in\ [Z_{min}, Z_{max}], e\in\ [0,255]$ respectively for source data. The depth label map is distributed in fixed intervals. Input image data will pass through the pretrained feature extractor ${\ F}_b$ and decoder $g$, to get predicted value $\hat{y_s}=\ \ g_s(F_b(x))$ for semantics, $\hat{z}=\ \ g_z(F_b(x))\ $ for depth, and $\hat{e}=\ \ g_e\left(F_b\left(x\right)\right)$ for edge. refined prediction value $\hat{y_r}=\ \ g_r 
 (\hat{z} \odot F_b(x))\ $ are employed to reflect semantics and depth. Each predicted feature map will be upscaled to the original size of the image. Semantic and depth information will be trained in supervised learning using cross-entropy loss and berHu loss, respectively, as follows:

\begin{equation}
\mathcal{L}_{seg}\left({\hat{y}}_s,y\right)=\ -\sum_{h=1}^{H}\sum_{w=1}^{W}\sum_{c=1}^{C}y_i^{\left(h,w,c\right)}logP{y_i}^{\left(h,w,c\right)},
\end{equation}

\begin{equation}
\mathcal{L}_{seg}\left({\hat{y}}_r,y\right)=\ -\sum_{h=1}^{H}\sum_{w=1}^{W}\sum_{c=1}^{C}y_i^{\left(h,w,c\right)}logP{y_i}^{\left(h,w,c\right)},
\end{equation}

\begin{equation}
\mathcal{L}_{dep}\left({\hat{z}}_s,z_s\right)=\ -\sum_{h=1}^{H}\sum_{w=1}^{W}{berHu(}z_{x_s}^{\left(h,w\right)}-z_s^{\left(h,w\right)}).
\end{equation}

The entropy probability can be obtained from the pixel prediction value for each semantic class. Eliminating uncertainty through the minimization of entropy probability can enhance adaptation performance.  The following formula can obtain the entropy probability:

\begin{equation}
\mathcal{E}(p)=\ -P_x^{\left(h,w,c\right)}\log{P_x^{\left(h,w,c\right)}}.
\end{equation}

Before calculating the entropy of depth, discretization is applied to $\hat{z}$ based on the SID \citep{fu2018deep} and CTRL \citep{saha2021learning} schemes. The final predicted entropy map follows the following:

\begin{equation}
E_s=\mathcal{E}\left(\hat{y_s}\right),\ \ 
E_r=\mathcal{E}\left(\hat{y_r}\right),\ \
E_z=\ \mathcal{E}(\hat{z}).
\end{equation}

\subsection{Concatenating Edge Probability}
The prediction entropy maps derived from scenes within the source domain resemble edge detection outcomes, displaying high entropy activations solely along the object borders~\citep{vu2019advent}. However, high entropy can be considered object boundaries due to its uncertainty rather than being the edge itself. Therefore, we explore enhancing the entropy of object boundaries utilizing edge probability. Training the edges from the source domain and applying them to the target domain makes it possible to attain high-quality edge probabilities without noise. For the training, we apply Canny edge detection to semantic labels to obtain edge labels. From the decoder of model, we obtained  $\hat{e}=\ \ g_e\left(F_b\left(x\right)\right)$. 
Then, we use sigmoid activation function $\sigma$ to predict final output $\hat{e}$ for normalized each pixel value as follows: 
 
\begin{equation}
{\widetilde{e}}_i=\ \ \sigma(\hat{e}).
\end{equation}

The edge prediction value \ $\widetilde{e}$ and ground truth $e$ are represented by discrete numbers 0 and 255, we use binary cross entropy loss for training as follows: 

\begin{equation}
\mathcal{L}_{BCE}({\widetilde{e}}_i, e_i)=\ -\sum_{h=1}^{H}\sum_{w=1}^{W}{}e_i \times log{\widetilde{e}}_{i} +(1-e_i ) \times\ log(1-{\widetilde{e}}_{i,}).
\end{equation}

To simultaneously consider the entropy of internal information and the external edge information, the fusion process is imperative. To enable the discriminator to account for the entropy and edges of both the source and target domains, prediction maps were combined. We build concatenate entropy map $E=concat[E_s, E_z, E_s]$ for internal information, and then finally concatenate semantic information and edge, $\widetilde{I}\ =\ concat[E,\widetilde{e}]$. 

\subsection{Discriminator}
Given ${\widetilde{I}}_s, {\widetilde{I}}_t$ the set of unified probability maps in each domain, the discriminator $D$ is trained to differentiate source and target. The discriminators can distinguish between real and fake by 0 or 1 extracted from each domain by reducing the domain gap until each domain value is indistinguishable. The discriminator is following below formula:

\begin{equation}
\mathcal\min{{}\mathbb{E}_{\mathcal{D}^{\left({s}\right)}}[logD(\widetilde{I}^{\left(s\right)})]+\mathbb{E}_{\mathcal{D}^{\left({t}\right)}}[1-logD(\widetilde{I}^{\left(t\right)})]}.
\end{equation}
    
Meanwhile, the discriminator understands the gap of domains, and the segmentation prediction network is updated to maximize classification loss to fool the discriminator: 

\begin{equation}
\mathcal\min_{\theta dis.}{\ \mathbb{E}}_{\mathcal{D}^{\left(t\right)}}[logD(\widetilde{I}^{\left(t\right)})].
\end{equation}

\subsection{Self-training}
 We finally used a self-training method that allows our model to train with pseudo-labels to unlabeled target data. A powerful domain adaptation technique, self-training can learn superior decision boundaries and find similar alignment of source and target domain distributions. For semantic segmentation, self-training methods aim to achieve the same output as adversarial network training access in alleviating the domain gap between source and target. The procedure was carried out as follows: First, supervised learning is performed with labeled data to obtain a fine classifier. We then predict the unlabeled target domain using the trained classifier. Lastly, only the data with the highest probability for each pixel were generated as pseudo-labels so that self-learning could be performed. We followed by CTRL \citep{saha2021learning} self-training scheme.

\section{Experiments}
\subsection{Dataset Overview}
To assess our proposed model for SS, we utilized a challenging synthetic-to-real task, specifically the \textbf{SYNTHIA} $\rightarrow$ \textbf{Cityscapes} and \textbf{SYNTHIA} $\rightarrow$ \textbf{Mapillary} scenarios. These experimental environments involve applying our approach to a virtual environment dataset and testing it in a real-world domain, allowing us to assess our method's generalizability and robustness.

In this experiment, we used the \textbf{SYNTHIA} dataset \citep{ros2016synthia} as the source domain, which serves as a virtual environment for urban scenes. This dataset consists of 9,400 RGB images automatically annotated with 12 class categories. In addition, the dataset includes annotated depth labels with a spatial resolution of 760 $\times$ 1280. We  applied Canny edge detection to the semantic ground truth to obtain edge labels. The \textbf{Cityscapes} dataset \citep{cordts2016cityscapes} comprises images captured in 50 different cities worldwide. This dataset includes 2,495 images for training and 500 images for testing, making it a valuable resource for evaluating the performance of semantic segmentation models in real-world urban environments. The \textbf{Mapillary} dataset \citep{neuhold2017mapillary} is another valuable resource for evaluating the performance of semantic segmentation models in real-world settings. This dataset contains 24,966 high-resolution images with 66 object categories. We used both the Cityscapes and Mapillary datasets as target domains in our experiments, allowing us to assess the generalizability of our proposed method across a range of real-world environments.

\subsection{Implementation Setup}
Our proposed method was implemented using the PyTorch framework on an NVIDIA RTX A6000 GPU. As our backbone network, we used a DeeplabV2~\citep{chen2017deeplab} network, which was pre-trained on the ImageNet dataset. Adversarial training was performed using a DC-GAN discriminator backbone \citep{radford2015unsupervised} consisting of 4 convolutional layers with a Leaky-ReLU activation function. We optimized each prediction task using SGD and the discriminator networks using Adam with learning rates of $2.5 \times 10^{-4}$ and $10^{-4}$, respectively. To generate depth features, we used $Z_{min} = 1m$ and $Z_{max} = 666.36m$, following the scheme proposed in \cite{saha2021learning}. 
\subsection{Results}
We compare our proposed method to several state-of-the-art unsupervised domain adaptation models for semantic segmentation, namely OutputAdapt \citep{tsai2018learning}, ADVENT \citep{vu2019advent}, CBST \citep{zou2018unsupervised}, SA-I2I \citep{musto2020semantically}, PIT \citep{lv2020cross}, IntraDA \citep{pan2020unsupervised}, 

LSE \citep{subhani2020learning}, ASA~\citep{zhou2020affinity}, FADA~\citep{wang2020classes}, MetaCorrect~\citep{guo2021metacorrection}, DAST~\citep{yu2021dast}, CIRN~\citep{gao2021addressing}, CCD~\citep{chen2021enhanced}, 
UDAClu \citep{toldo2021unsupervised}, 
CRST \cite{kong2022constraining},
\cite{li2022feature},
\cite{zhang2022confidence},
CBNA \cite{klingner2022continual},
ProCA \citep{jiang2022prototypical}, ARAS \citep{cao2023adaptive}, SPIGAN \citep{lee2018spigan}, GIO-Ada \citep{chen2019learning}, DADA \citep{vu2019dada}, and CTRL\citep{saha2021learning}. We evaluate our approach on the SYNTHIA $\rightarrow$ Cityscapes task, where each semantic is divided into 16 classes, and The mean intersection over union (mIoU\%) is utilized as an assessment criterion. Our approach achieves the highest performance compared to the previous methods.
\renewcommand{\arraystretch}{1.8}
\begin{landscape}
\begin{table}[!ht]
\centering
\caption{Quantitative comparison of model performance based on SYNTHIA-trained semantic segmentation models on Cityscapes. The top and bottom sub-tables are divided to represent whether depth prediction value is used or not. Our edge probability concatenation approach provides the best mIoU results.}
\resizebox{\textwidth}{!}{%
\label{t1_16cls}
\begin{tabular}{c|c|cccccccccccccccc|c}
\hline
\multicolumn{19}{c}{\LARGE{SYNTHIA $\rightarrow$ Cityscapes}} \\
\hline
& Dep. & road & s.walk & build. & wall & fence & pole & light & sign & veg. & sky & person & rider & car & bus & mbike & bike & mIoU \\
\hline
OutputAdapt \cite{tsai2018learning} &  & 84.3 & 42.7 & 320 & - & - & - & 4.7 & 7.0 & 77.9 & 82.5 & 54.3 & 21.0 & 72.3 & 32.2 & 18.9 & 32.3 & - \\
ADVENT \cite{vu2019advent} & & 85.6 & 42.2 & 79.7 & 8.7 & 0.4 & 25.9 & 5.4 & 8.1 & 80.4 & 84.1 & 57.9 & 23.8 & 73.3 & 36.4 & 14.2 & 33.0 & 41.2 \\
CBST \cite{zou2018unsupervised} &   & 68.0 & 29.9 & 76.3 & 10.8 & 1.4 & 33.9 & 22.8 & 29.5 & 77.6 & 78.3 & 60.6 & 28.3 & 81.6 & 23.5 & 18.8 & 39.8 & 42.6 \\
SA-I2I \cite{musto2020semantically} &   & 79.1 & 34.0 & 78.3 &0.3 & 0.6 & 26.7 & 15.9 & 29.5 & 81.0 & 81.1 & 55.5 & 21.9 & 77.2 & 23.5 & 11.8 & 47.5 & 41.5 \\
PIT \cite{lv2020cross} &   & 73.7 & 29.6 & 77.6 & 1.0 & 0.4 & 26.0 & 14.7 & 26.6 & 80.6 & 81.8 & 57.2 & 24.5 & 76.1 & 27.6 & 13.6 & 46.6 & 41.1 \\
IntraDA \cite{pan2020unsupervised} &   & 84.3 & 37.7 & 79.5 & 5.3 & 0.4 & 24.9 & 9.2 & 8.4 & 80.0 & 84.1 & 57.2 & 23.0 & 78.0 & 38.1 & 20.3 & 36.5 & 41.7 \\
LSE \cite{subhani2020learning} &  & 82.9 & 43.1 & 78.1 & 9.3 & 0.6 & 28.2 & 9.1 & 14.4 & 77.0 & 83.5 & 58.1 & 25.9 & 71.9 & 38.0 & \textbf{29.4} & 31.2 & 42.6 \\ 
ASA \cite{zhou2020affinity} &  & 91.2 & \textbf{48.5} & 80.4 & 3.7 & 0.3 & 21.7 & 5.5 & 5.2 & 79.5 & 83.6 & 56.4 & 21.0 & 80.3 & 36.2 & 20.0 & 32.9 & 41.7 \\
FADA \cite{wang2020classes} &  & 84.5 & 40.1 & \textbf{83.1} & 4.8 & 0.0 & 34.3 & 20.1 & 27.2 & \textbf{84.8} & 84.0 & 53.5 & 22.6 & \textbf{85.4} & 43.7 & 26.8 & 27.8 & 45.2 \\
MetaCorrect \cite{guo2021metacorrection} &  & \textbf{92.6} &  52.7 & 81.3 & 8.9 & 2.4 & 28.1 & 13.0 & 7.3 & 83.5 & 85.0 & 60.1 & 19.7 & 84.8 & 37.2 & 21.5 & 43.9 & 45.1 \\ 
DAST \cite{yu2021dast} &  & 87.1 & 44.5 & 82.3 & 10.7 & 0.8 & 29.9 & 13.9 & 13.1 & 81.6 & 86.0 & 60.3 & 25.1 & 83.1 & 40.1 & 24.4 & 40.5 & 45.2 \\
CIRN \cite{gao2021addressing} &  & 85.8 & 40.4 & 80.4 & 4.7 & 1.8 & 30.8 & 16.4 & 18.6 & 80.7 & 80.4 & 55.2 & 26.3 & 83.9 & 43.8 & 18.6 & 34.3 & 43.9 \\
CCD \cite{chen2021enhanced} &  & 74.4 & 28.8 & 81.5 & 13.5 & 1.2 & 32.6 & 21.6 & 32.4 & 81.5 & 83.7 & 52.8 & 25.8 & 78.0 & 30.0 & 29.6 & 52.7 & 45.0 \\
UDAClu \cite{toldo2021unsupervised} &  & 64.4 & 25.5 & 77.3 & 14.3 & 0.9 & 29.6 & 21.2 & 24.2 & 76.6 & 79.7 & 53.7 & 15.5 & 79.7 & 11.0 & 11.0 & 35.2 & 38.7 \\
CRST \cite{kong2022constraining} &  & 67.6 & 24.9 & 78.9 & 18.0 & 9.2 & 36.9 & 26.9 & 35.5 & 82.4 & 79.4 & 59.5 & 29.8 & 85.0 & 29.9 & 22.0 & 47.4 & 45.7 \\
FRA-RPLR\cite{li2022feature} &  & 81.5 & 36.7 & 78.6 & 1.3 & 0.9 & 32.2 & 20.7 & 23.6 & 79.1 & 83.4 & 57.6 & \textbf{30.4} & 78.5 & 38.3 & 24.7 & \textbf{48.4} & 44.7 \\
CRAM(+BDL)\cite{zhang2022confidence} &  & 87.6 & 46.1 & 82.0 & 10.0 & 0.4 & 33.6 & 21.4 & 14.9 & 81.2 & 85.2 & 57.2 & 26.4 & 83.0 & 33.3 & 24.0 & 46.8 & 45.8 \\
CBNA \cite{klingner2022continual} &  & 79.9 & 46.7 & 74.5 & 10.5 & \textbf{10.2} & \textbf{41.3} & \textbf{28.3} & \textbf{39.1} & 84.3 & \textbf{88.6} & 59.0 & 18.5 & 75.8 & 14.5 & 4.1 & 37.0 & 44.5\\
ProCA \cite{jiang2022prototypical} &  & 81.8 & 38.6 & 80.3 & 9.5 & 8.1 & 28.7 & 16.6 & 16.0 & 80.6 & 78.6 & 55.8 & 21.2 & 79.7 & 31.6 & 14.3 & 41.3 & 42.2 \\
ARAS \cite{cao2023adaptive} &  & 85.6 & 39.2 & 79.9 & 15.5 & 0.3 & 32.2 & 19.3 & 23.9 & 79.1 & 81.7 & 61.1 & 19.3 & 82.9 & 25.7 & 10.6 & 51.9 & 44.3 \\

\hline 
SPIGAN \cite{lee2018spigan} & \checkmark &   71.1 & 29.8 & 71.4 & 3.7 & 0.3 & 33.2 & 6.4 & 15.6 & 81.2 & 78.9 & 52.7 & 13.1 & 75.9 & 25.5 & 10.0 & 20.5 & 36.8\\
GIO-Ada \cite{chen2019learning} & \checkmark &   78.3 & 29.8 & 71.4 & 3.7 & 0.3 & 33.2 & 6.4 & 15.6 & 81.2 & 78.9 & 52.7 & 13.1 & 75.9 & 25.5 & 10.0 & 20.5 & 36.8 \\
DADA \cite{vu2019dada} & \checkmark &   89.2 & 44.8 & 81.4 & 6.8 & 0.3 & 26.2 & 8.6 & 11.1 & 81.8 & 84.0 & 54.7 & 19.3 & 79.7 & 40.7 & 14.0 & 38.8 & 42.6 \\
CTRL \cite{saha2021learning} & \checkmark &   86.4 & 42.5 & 80.4 & \textbf{20.0} & 1.0 & 27.7 & 10.5 & 13.3 & 80.6 & 82.6 & 61.0 & 23.7 & 81.8 & 42.9 & 21.0 & 44.7 & 45.0 \\
\hline
\hline
Ours & \checkmark  & 88.5 & 42.7 & 82.9 & 14.9 & 0.4 & 28.7 & 10.6 & 11 & 83.8 & 85.8 & \textbf{63.6} & 27.3 & 73.0 & \textbf{48.6} & 28.9 & 47.0 & \textbf{46.1} \\
\hline
\hline
\end{tabular}%
}
\end{table}
\end{landscape}

\begin{figure}[!t]
\centering
\includegraphics[width=0.9\linewidth]{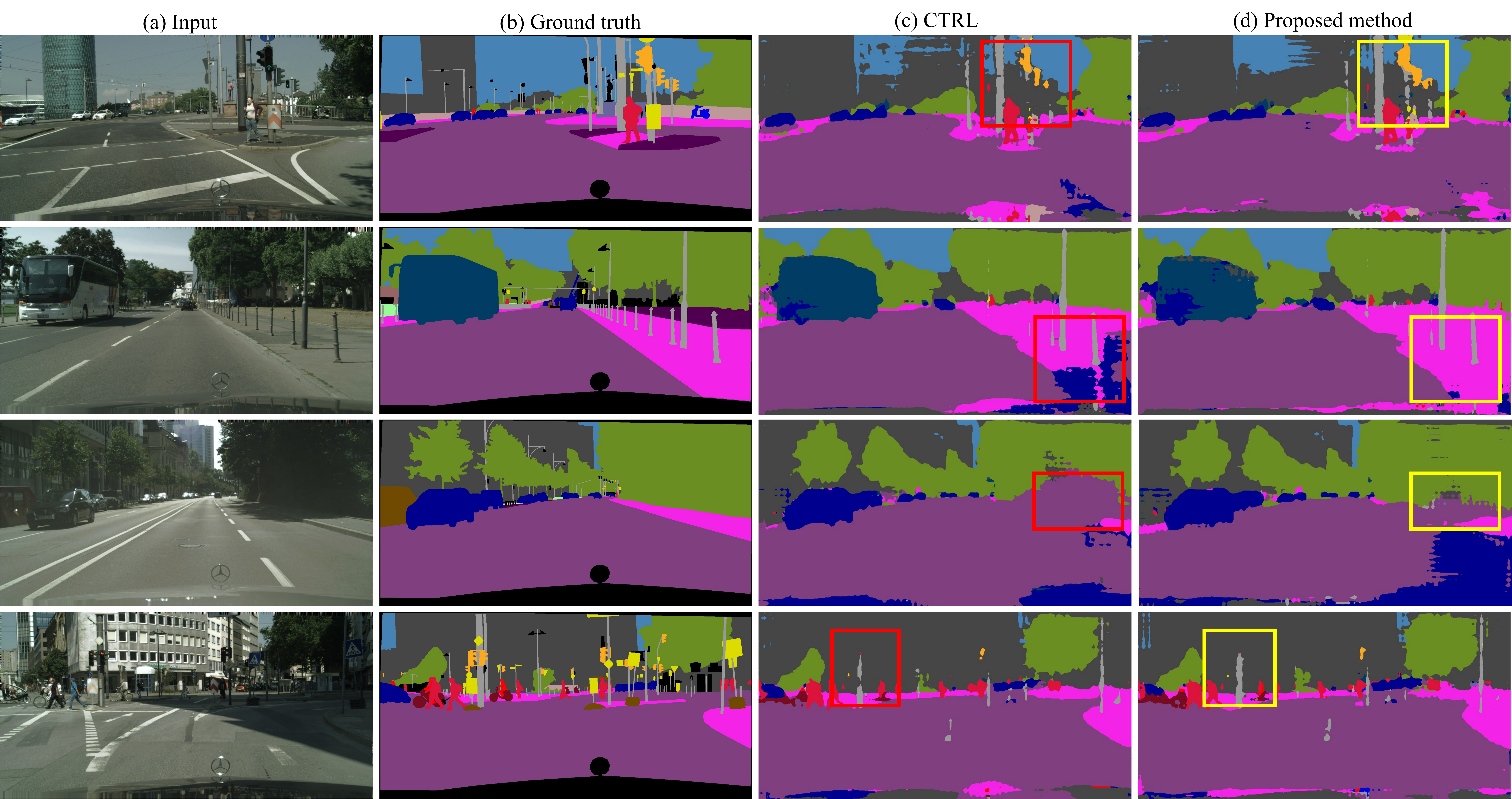}
\caption{Qualitative semantic segmentation results with SYNTHIA $\rightarrow$ Cityscapes (16 classes). The first column is an RGB image; the second column is the ground truth; the third column is the results of the CTRL method; the fourth column is the results of our proposed method result. Our model is robust at the boundary of each semantic.}
\label{visual_result}
\end{figure}

Table \ref{t1_16cls} shows the qualitative results of unsupervised domain adaptation in semantic segmentation performance. Each experiment takes around 16 hours for a 16-class prediction task. We divide the results into two sections: methods without depth and methods with depth. Recent depth techniques such as DADA \citep{vu2019dada} and CTRL \citep{saha2021learning} have demonstrated competitive performance, particularly for large objects like buildings, buses, skies, vegetation, and sidewalks. However, our proposed method outperforms these methods by considering edge probabilities with depth information. Specifically, our approach shows superior performance compared to previous methods in several classes such as a person (+2.6\%), vegetation (+3.2\%), sky (+3.2\%), bike (+2.3\%), and building (+2.5\%). Our analysis identifies significant improvements in pixel-specific features over the state-of-the-art CTRL (+1.1\%). In certain classes that include features such as fences and lighting, outstanding performance could not be achieved. These particular classes presented a unique set of complexities that limited the model's performance potential. One noteworthy factor contributing to these challenges is the demand for highly localized object detection within the original image. In the case of features like fences and lighting, achieving precise localization becomes crucial for accurate identification. However, this precise localization often encounters obstacles in the form of potential confusion with adjacent pixel labels. For instance, in scenarios where fences are prevalent, the model may struggle to distinguish between the fine details of the fence structure and background elements, leading to occasional misclassifications or suboptimal results. Similarly, when dealing with lighting features, the model's performance may be affected by the intricate interplay of light and shadow, making it more challenging to achieve consistently outstanding detection accuracy.

Figure \ref{visual_result} displays the qualitative results of the proposed method on test and validation samples of Cityscape datasets. Our approach generates fine segmentation output with a good representation of context details near the boundaries. Comparing the yellow and red boxes, yellow boxes indicate improved results that can be perceived intuitively. Moreover, for pole and light interfaces at rows 1 and 4, our proposed method outperforms the previous method in terms of robustness by segmenting accurately. Similarly, for boundaries near the road and vegetation at rows 2 and 3, our proposed method exhibits a higher capability of preserving pixel prediction along each class prediction. These results demonstrate the significance of our method in generating accurate semantic segmentation maps.

\begin{table}[!ht]
\centering
\caption{Semantic segmentation results comparison to previous methods for the SYNTHIA $\rightarrow$ Cityscapes (7 classes).}
\resizebox{\columnwidth}{!}{%
\label{t2_city_7cls}
\begin{tabular}{c|c|c|ccccccc|c}
\hline
\multicolumn{11}{c}{SYNTHIA $\rightarrow$ Cityscapes (7 classes)} \\
\hline
%
& Dep. & Edge & flat & const. & object & nature & sky & human & vehicle & mIoU \\
\hline
ADVENT \citep{vu2019advent} &  &  & 89.6 & 77.8 & 22.1 & 76.3 & 81.4 & 54.7 & 68.7 & 67.2 \\
DADA \citep{vu2019dada} & \checkmark & & 92.3 & 78.3 & 25.0 & 75.5 & 82.2 & 58.7 & 72.4 & 69.2 \\
CTRL \citep{saha2021learning} & \checkmark & & 91.1 & 79.8 & 27.5 & 73.9 & 83.3 & 62.1 & 75.8 & 70.5 \\
\hline
\hline
Ours & \checkmark & \checkmark & \textbf{92.7} & \textbf{80.1} & \textbf{28.0} & \textbf{76.8} & \textbf{83.9} & \textbf{63.3} & \textbf{79.4} & \textbf{72.0} \\
\hline
\hline
\end{tabular}%
}
\end{table}

\renewcommand{\arraystretch}{1.5}
\begin{table}[!ht]
\centering
\caption{Comparison of semantic segmentation results to previous methods for the SYNTHIA $\rightarrow$ 
 Mapillary (7 classes).}
\resizebox{\columnwidth}{!}{%
\label{t4_map_7cls}
\begin{tabular}{c|c|c|ccccccc|c}
\hline
\multicolumn{11}{c}{SYNTHIA $\rightarrow$ Mapillary (7 classes)} \\
\hline
%
& Dep. & Edge & flat & const. & object & nature & sky & human & vehicle & mIoU \\
\hline
ADVENT \citep{vu2019advent} &  &  & 86.9 & 58.8 & 30.5 & 74.1 & 85.1 & 48.3 & 72.5 & 65.2 \\
DADA \citep{vu2019dada} & \checkmark & & 86.7 & \textbf{62.1} & 34.9 & 75.9 & \textbf{88.6} & 51.1 & 73.8 & 67.6\\
CTRL \citep{saha2021learning} & \checkmark & & 86.5 & 59.3 & \textbf{35.3} & 76.0 & 85.8 & 61.6 & 77.9 & 68.9\\
\hline
\hline
Ours & \checkmark & \checkmark & \textbf{88.5} & 59.4 & 32.4 & \textbf{79.1} & 86.2 & \textbf{63.2} & \textbf{78.5} & \textbf{69.6} \\
\hline
\hline
\end{tabular}%
}
\end{table}
\label{table:uavidtable}

The validation results for the SYNTHIA $\rightarrow$  Cityscape and SYNTHIA $\rightarrow$  Mapillary tasks for seven classes are presented in Table \ref{t2_city_7cls} and \ref{t4_map_7cls}. For comparison, we selected ADVENT \citep{vu2019advent}, DADA \citep{vu2019dada}, and CTRL \citep{saha2021learning}. Table \ref{t2_city_7cls} shows the results of the same experimental setup as Table \ref{t1_16cls}, except that training and validation were performed on seven categories. The seven classes were obtained by merging 16 categories: flat, construction, object, nature, sky, human, and vehicle. Our proposed edge-combined framework remarkably surpasses the state-of-the-art on this benchmark task, with improvements shown in `flat' (+1.6\%), `construction' (+0.3\%), `nature' (+2.3\%), `object' (+0.5\%), `nature' (+2.8\%), `sky' (+0.6\%), `human' (+1.2\%), `vehicle' (+3.6\%), and mIoU (+1.5\%). Furthermore, we trained and evaluated our model on the Mapillary benchmark dataset for 7 classes in Table \ref{t4_map_7cls}. Our approach achieved the highest performance in mIoU (0.7+\%) compared to CTRL, especially in `flat' (+2.0\%), `nature' (+3.1\%), `human' (+1.6\%), and `vehicle' (+0.6\%).


\renewcommand{\arraystretch}{1.5}
\begin{table}[!ht]
\centering
\caption{Semantic segmentation results comparison to previous methods for the SYNTHIA $\rightarrow$ 
 Cityscapes (7 classes). Low resolution is consists of 320 $\times$ 640. We trained semantic segmentation at a low resolution.}
\resizebox{\columnwidth}{!}{%
\label{t3_city_7clslr}
\begin{tabular}{c|c|c|ccccccc|c}
\hline
\multicolumn{11}{c}{SYHTNIA $\rightarrow$ Cityscapes (7 classes on low-resolution)} \\
\hline
%
& Dep. & Edge & flat & const. & object & nature & sky & human & vehicle & mIoU \\
\hline
SPIGAN-no-PI \citep{lee2018spigan} & & & 90.3 & 58.2 & 6.8 & 35.8 & 69.0 & 9.5 & 52.1 & 46.0 \\
SPIGAN \citep{lee2018spigan} & \checkmark & & 91.2 & 66.4 & 9.6 & 56.8 & 71.5 & 17.7 & 60.3 & 53.4 \\   
ADVENT \citep{vu2019advent} &  &  & 86.3 & 72.7 & 12.0 & 70.4 & 81.2 & 29.8 & 62.9 & 59.4 \\
DADA \citep{vu2019dada} & \checkmark & & 89.6 & 76.0 & 16.3 & 74.4 & 78.3 & 43.8 & 65.7 & 63.4 \\
CTRL \citep{saha2021learning} & \checkmark & & 90.8 & \textbf{77.5} & 15.7 & \textbf{77.1} & 82.9 & 45.3 & 68.6 & 65.4 \\
\hline
\hline
Ours & \checkmark & \checkmark & \textbf{91.5} & 77.0 & \textbf{17.5} & 75.9 & \textbf{84.1} & \textbf{45.8} & \textbf{69.6} & \textbf{66.0} \\
\hline
\hline
\end{tabular}%
}
\label{tablel:psi}
\end{table}

\renewcommand{\arraystretch}{1.5}
\begin{table}[!ht]
\centering
\caption{Comparison of semantic segmentation results to previous methods for the SYNTHIA $\rightarrow$ 
 Mapillary (7 classes). Low resolution is consists of 320 $\times$ 640. We trained semantic segmentation at a low resolution.}
\resizebox{\columnwidth}{!}{%
\label{t5_map_7clslr}
\begin{tabular}{c|c|c|ccccccc|c}
\hline
\multicolumn{11}{c}{SYNTHIA $\rightarrow$  Mapillary (7 classes on low-resolution)} \\
\hline
%
& Dep. & Edge & flat & const. & object & nature & sky & human & vehicle & mIoU \\
\hline
SPIGAN-no-PI \citep{lee2018spigan} & & & 53.0 & 30.8 & 3.6 & 14.6 & 53.0 & 5.8 & 26.9 & 26.8 \\
SPIGAN \citep{lee2018spigan} & \checkmark & & 74.1 & 47.1 & 6.8 & 43.3 & 83.7 & 11.2 & 42.2 & 44.1 \\
ADVENT \citep{vu2019advent} &  &  & 82.7 & 51.8 & 18.4 & 67.8 & 79.5 & 22.7 & 54.9 & 54.0 \\
DADA \citep{vu2019dada} & \checkmark & & 83.8 & 53.7 & 20.5 & 62.1 & 84.5 & 26.6 & 59.2 & 55.8 \\
CTRL \citep{saha2021learning} & \checkmark & & \textbf{86.6} & \textbf{57.4} & 19.7 & 73.0 & \textbf{87.5} & 45.1 & \textbf{68.1} & 62.5 \\
\hline
\hline
Ours & \checkmark & \checkmark & 84.7 & 55.0 & \textbf{20.4} & \textbf{73.3} & 86.8 & \textbf{53.1} & 67.0 & \textbf{62.9} \\
\hline
\hline
\end{tabular}%
}
\end{table}

Table \ref{t3_city_7clslr} and \ref{t5_map_7clslr} evaluate seven classes similar to the previous experiment but with a lower resolution (320 $\times$ 640). The original size of the high-resolution training input size is (760 $\times$ 1280) for the Cityscapes dataset and (768 $\times$ 1024) for the Mapillary dataset, respectively. To compare the results, we selected SPIGAN \citep{lee2018spigan}, ADVENT \citep{vu2019advent}, DADA \citep{vu2019dada}, and CTRL \citep{saha2021learning}. Learning at low-resolution results in lower performance than experiments at full resolution due to the limited amount of information available. When comparing

Table \ref{t2_city_7cls} and \ref{t3_city_7clslr} reported that the mIoU performance learned at low resolution decreased compared to that learned at full resolution. Especially, ``human”, ``objects” and ``vehicles” showed a significant decrease. Nevertheless, the proposed technique acquired a better mIoU score on the SYNTHIA $\rightarrow$ Cityscape task in low-resolution experiments compared to the previous method, with improvement in most classes, including ‘flat’ (+0.7\%), ‘object’ (+1.8\%), ‘sky’ (+1.2\%), ‘human’ (+0.5\%), ‘vehicle’ (+1\%) and mIoU (+0.6\%). Similar results can be seen in Table 5 for the SYNTHIA $\rightarrow$ Mapillary task. Our method showed better performance in some classes, with a (+0.4\%) higher mIoU performance than CTRL. ‘object’ (+0.7\%), ‘nature (+0.3\%)’ and ‘human’ (+8\%) classes showed superior performance than previous methods. 

While our proposed algorithm excels in many aspects of object detection, it's important to acknowledge certain limitations, particularly in classes that encompass intricate features like fences and lighting conditions. In these cases, achieving outstanding performance has proven to be a challenge. One of the key limitations stems from the inherent nature of highly localized detection required in the original image. When dealing with such fine-grained details, the algorithm may encounter difficulties in distinguishing between subtle variations in pixel labels, which can lead to potential confusion.
For instance, when detecting objects like fences, which often consist of thin, closely spaced lines, or objects under varying lighting conditions that create subtle gradients and shadows, the algorithm may struggle to precisely delineate boundaries. This limitation arises due to the intricate interplay between pixel values, making it challenging to differentiate between foreground and background accurately. Furthermore, these limitations are exacerbated when objects of interest share visual characteristics with other pixel labels in the image. In such cases, the algorithm's performance may be hindered as it grapples with distinguishing between the target objects and the background or similar objects. To address these challenges and further improve the algorithm's robustness, future work could explore techniques for enhancing the detection of fine-grained features, as well as strategies to mitigate potential confusion in highly localized detection scenarios. Despite these limitations, our algorithm demonstrates promising results in various other object detection scenarios, showcasing its versatility and potential for further refinement.

\begin{figure}[!t]
\centering
\includegraphics[width=0.9\linewidth]{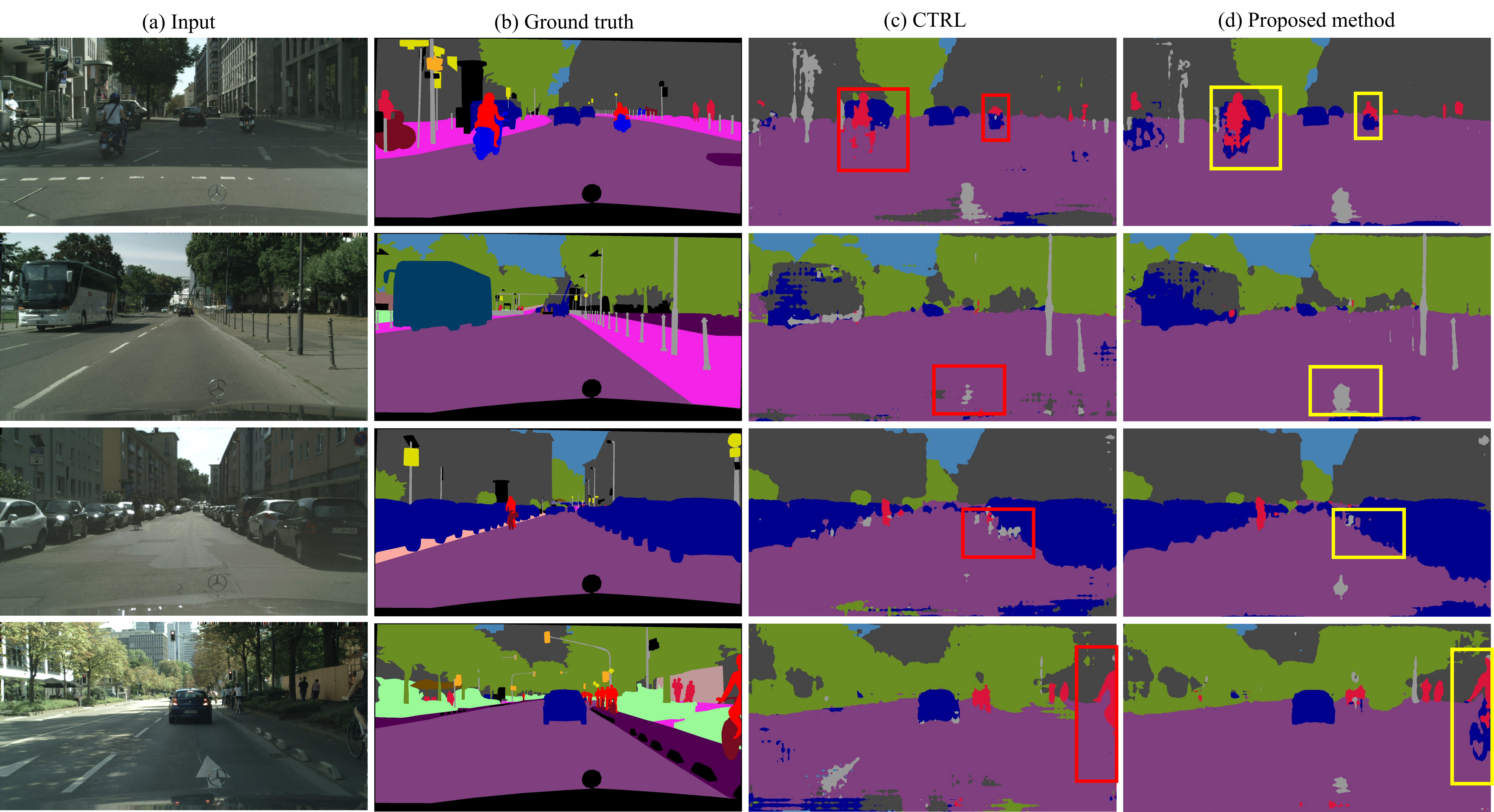}
\caption{Qualitative semantic segmentation results with SYNTHIA $\rightarrow$ Cityscapes (7 classes). The first column is an RGB image; the second column is the ground truth for semantics; the third column is the results of the CTRL method; the fourth column is the results of our proposed method. Our model is robust at the boundary of each semantic.}
\label{visual_result2}
\end{figure}

\begin{figure}[!t]
\centering
\includegraphics[width=0.9\linewidth]{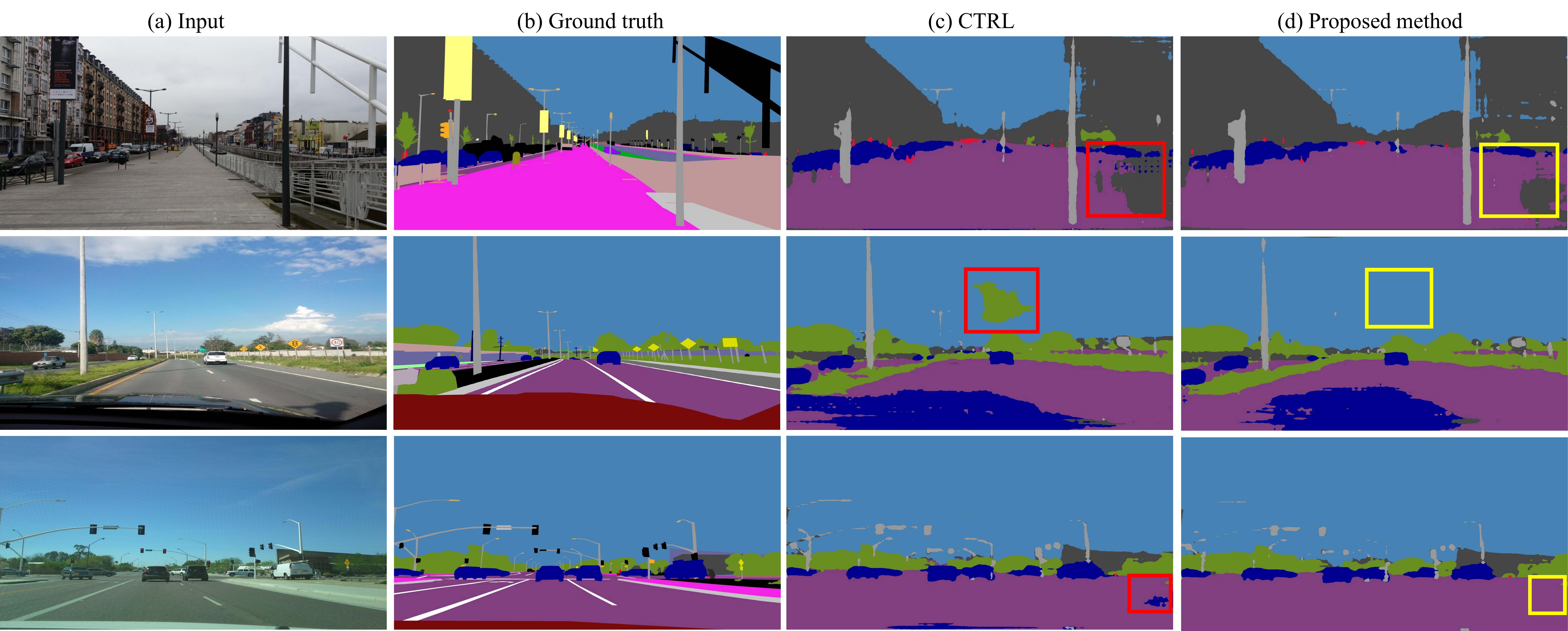}
\caption{Qualitative semantic segmentation with SYNTHIA $\rightarrow$ Mapillary (7 classes). The first column is an RGB image; the second column is the ground truth for semantics; the third column is the results of the CTRL method; the fourth column is the results of our proposed method. Our model is robust at the boundary of each semantic.}
\label{visual_result3}
\end{figure}

\FloatBarrier
\subsection{Ablative studies}
To gain a deeper understanding of our model's effectiveness and its various components, we conduct a thorough ablation study. This study thoroughly examines the impact of using entropy and edge probabilities as inputs for the discriminator to perform adversarial training. By carefully assessing these components individually, we aim to uncover how they influence our model's overall performance. 
\subsubsection{Impact of hyperparameter $\lambda_{\text{conf.}}$}
In this experimental study, we conducted a comprehensive ablation analysis aimed at fine-tuning the performance of our model by carefully adjusting the confidence threshold hyperparameter associated with the self-training process. The concept of confidence analysis within the context of self-training draws inspiration from earlier research by ~\cite{saha2021learning}. Specifically, we employed a crucial hyperparameter denoted as $\lambda_{\text{conf.}}$, which is pivotal in shaping the self-training process. This hyperparameter is set to a value of 0.8, following the methodology established in prior work. Traditionally, self-training methods typically utilized a confidence threshold of 0.9 to determine the reliability of predictions for target pixels. However, our study observed a shift in the network's sensitivity to this hyperparameter with the integration of edges into the entropy computation. To comprehensively evaluate the impact of varying confidence thresholds, we conducted a series of experiments, systematically varying the confidence threshold within the range from 0.6 to 0.8.

The results of our ablation study, presented in Table \ref{Ablative(confidence)}, clearly demonstrate the significance of this adjustment. Specifically, we observe that setting the confidence threshold to 0.8 yielded a substantial improvement in the self-training process, as evidenced by the notable increase in mIoU. This empirical finding underscores the critical role of the confidence threshold hyperparameter in optimizing the performance of our model within the domain adaptation context.

\renewcommand{\arraystretch}{1.5}
\begin{table}[!ht]
\centering
  \caption{The parameter sensitivities of confidence pseudo label for the self-training to perform the SYNTHIA $\rightarrow$ Cityscapes task.}
\resizebox{0.3\linewidth}{!}{%
\label{Ablative(confidence)}
\begin{tabular}{ c c c }
\hline
\multicolumn{3}{c}{SYNTHIA $\rightarrow$ Cityscapes} \\
\hline
 $\lambda_{conf.}$ & & mIoU \\
\hline
 $\lambda_{0.6}$ & & 45.80 \\
 $\lambda_{0.7}$ & & 45.95 \\
 $\lambda_{0.8}$ & & \textbf{46.10} \\
 $\lambda_{0.9}$ & & 46.08\\ 
\hline
\end{tabular}%
}
\end{table}

\renewcommand{\arraystretch}{1.5}
\begin{table}[!ht]
\centering
  \caption{Fusion analysis between edge and entropy for (a) SYNTHIA $\rightarrow$ Cityscapes and (b) SYNTHIA $\rightarrow$ Mapillary.}
\resizebox{0.6\linewidth}{!}{%
\label{Ablative(concatenation)}
\begin{tabular}{ c|c c c c c}
\hline
\multicolumn{6}{c}{ (a) SYNTHIA $\rightarrow$ Cityscapes} \\
\hline
 Methods & fusion & concat. & mIoU  & Param. & Flops(GMac) \\
\hline
 Entropy only &  &  & 45.00 & 48.6M & 164.1\\
 w. Edge & \checkmark &  & 44.45 & 48.6M & 164.1\\
 w. Edge to each entropy &  & \checkmark & 44.40 & 48.6M & 164.1\\
 w. Edge &  & \checkmark & \textbf{46.10} & 48.6M & 164.1\\ 
\hline
\end{tabular}%
}
\end{table}

\renewcommand{\arraystretch}{1.5}
\begin{table}[!ht]
\centering
\resizebox{0.6\linewidth}{!}{%
\begin{tabular}{ c|c c c c c}
\hline
\multicolumn{6}{c}{ (b) SYNTHIA $\rightarrow$ Mapillary} \\
\hline
 Methods & fusion & concat. & mIoU & Param. & Flops(GMac) \\
\hline
 Entropy only &  &  & 68.9 & 47.2M & 159.7\\
 w. Edge & \checkmark &  & 68.4 & 47.2M & 159.7\\
 w. Edge to each entropy & & \checkmark & 68.6 & 47.2M & 159.7\\
 w. Edge &  & \checkmark & \textbf{69.6} & 47.2M & 159.7\\  
\hline
\end{tabular}%
}
\end{table}

\subsubsection{Fusion analysis between edge and entropy}
In this experiment, we delve into the ablation study designed to provide a comprehensive understanding of the factors contributing to the observed performance improvements in our proposed method. We conduct analytical experiments that shed light on the underlying reasons for these enhancements, offering a deeper insight into the achieved performance. Our ablation study encompasses three distinct scenarios, each centered around using entropy maps in conjunction with different strategies: fusion and combination. In particular, we focus on elucidating the fusion strategy employed in our approach, as it plays a pivotal role in effectively combining entropy and edge information. The concept of entropy, in this context, signifies the degree of probability uncertainty associated with the extracted feature pixels. This uncertainty is analogous to the probability distribution of classes for individual pixels. While semantics and depth information allow for entropy calculation based on the number of classes, edges complicate the entropy calculation, as edges are typically trained with a single class. Given this challenge, our investigation reveals that the fusion strategy employing element-wise multiplication is unsuited for this specific task. Instead, a concatenation-based fusion strategy proves to be more suitable. This strategy preserves the values of probabilities and ensures that the information can be seamlessly transmitted to the discriminator.

To systematically explore the impact of these fusion strategies, we conduct experiments that fall into four distinct cases, as outlined in Table \ref{Ablative(concatenation)}. Here, `w' denotes `with,' and `w Edge to each entropy' represents an approach in which edge probability maps are combined with all three probability maps extracted.  Our experimental results, conducted on the SYNTHIA $\rightarrow$ Cityscapes dataset with 16 classes and the SYNTHIA $\rightarrow$ Mapillary dataset with seven classes, consistently demonstrate the highest performance achieved at 46.10\% and 69.60\%, respectively. These findings underscore the effectiveness of our proposed concatenation-based fusion strategy in enhancing the performance of our method.

\subsubsection{Quantifying the performance in terms of mean and standard deviation}
In addition to our comprehensive ablation study, we have further extended our evaluation by providing additional results in terms of the mean and standard deviation (Std.) of the mIoU metric. This supplementary analysis aims to offer a more comprehensive understanding of the stability and consistency of our proposed method across different experimental settings. By reporting both the mean and std of mIoU, we shed light on the overall performance and the variance in performance across multiple runs or scenarios.
\begin{table}
\centering
\caption{Statistical mean and standard deviation analysis for different experimental test cases.}
\label{mean}
\begin{tabular}{c|c|c|c c c }
\hline
Resultion & Episode & Number of class &Best mIoU & Mean & Std. \\
\hline
\multirow{3}{*}{\shortstack{Full \\ (760 $\times$ 1280)}} & SYNTHIA $\rightarrow$ Cityscapes & 16 & 46.1 & 45.8 & $\pm$ 0.3 \\
 & SYNTHIA $\rightarrow$ Cityscapes & 7 & 72.0 & 71.6 & $\pm$ 0.2 \\
 & SYNTHIA $\rightarrow$ Mapillary & 7  & 69.6 & 69.1 & $\pm$ 0.3 \\
\hline
\multirow{2}{*}{\shortstack{Low \\(320 $\times$ 640)}} &SYNTHIA $\rightarrow$ Cityscapes & 7 & 66.0  & 65.9 & $\pm$ 0.1 \\
& SYNTHIA $\rightarrow$ Mapillary & 7  & 62.9  & 62.7 & $\pm$ 0.2 \\ 
\hline
\end{tabular}%
\end{table}

The experimental results are presented in Table \ref{mean}. As the table indicates, every episode demonstrates a minimal standard deviation, falling within the range of 0.1 to 0.3. This collective data suggests that the model maintains a high degree of stability throughout all episodes, effectively mitigating concerns of underfitting or overfitting on the validation dataset. Nevertheless, it is worth noting that episodes sharing identical input image resolutions may exhibit slightly reduced model stability, particularly when a substantial volume of training data or classes is involved.

\section{Conclusion}
We introduce a novel approach to address the UDA task by reducing the domain gap between two distinct domains. Our method leverages internal and external semantic information to achieve this goal. Specifically, we utilize state-of-the-art techniques to obtain depth and semantic results in the first step. However, as no annotated labels are available for the edge detection task, we derive edge labels from the ground truth data. Finally, we concatenate predicted probabilities for each pixel from both the edge and semantic information to train the discriminator. By incorporating these two rich and complementary context information sources, our proposed network effectively combines entropy-based edges and semantic probabilities, which complement each other on a per-pixel basis. To demonstrate the efficacy of our approach, we evaluate it on synthetic-2-real scenarios and observe significant performance improvements compared to existing methods.

\section*{Acknowledgement}
This work was supported by the National Research Foundation of Korea (NRF) grant funded by the Korean government (MSIT) (No.2021R1C1C1012590) and (No.2022R1A4A1023248) 

\bibliography{sample}

\begin{thebibliography}{89}
\expandafter\ifx\csname natexlab\endcsname\relax\def\natexlab#1{#1}\fi
\providecommand{\url}[1]{\texttt{#1}}
\providecommand{\href}[2]{#2}
\providecommand{\path}[1]{#1}
\providecommand{\DOIprefix}{doi:}
\providecommand{\ArXivprefix}{arXiv:}
\providecommand{\URLprefix}{URL: }
\providecommand{\Pubmedprefix}{pmid:}
\providecommand{\doi}[1]{\href{http://dx.doi.org/#1}{\path{#1}}}
\providecommand{\Pubmed}[1]{\href{pmid:#1}{\path{#1}}}
\providecommand{\bibinfo}[2]{#2}
\ifx\xfnm\relax \def\xfnm[#1]{\unskip,\space#1}\fi
\bibitem[{Anand et~al.(2021)Anand, Sinha, Mandal, Chamola \& Yu}]{anand2021agrisegnet}
\bibinfo{author}{Anand, T.}, \bibinfo{author}{Sinha, S.}, \bibinfo{author}{Mandal, M.}, \bibinfo{author}{Chamola, V.}, \& \bibinfo{author}{Yu, F.~R.} (\bibinfo{year}{2021}).
\newblock \bibinfo{title}{Agrisegnet: Deep aerial semantic segmentation framework for iot-assisted precision agriculture}.
\newblock {\it \bibinfo{journal}{IEEE Sensors Journal}\/},  {\it \bibinfo{volume}{21}\/}, \bibinfo{pages}{17581--17590}. \DOIprefix\doi{https://doi.org/10.1109/JSEN.2021.3071290}.
\bibitem[{Asgari~Taghanaki et~al.(2021)Asgari~Taghanaki, Abhishek, Cohen, Cohen-Adad \& Hamarneh}]{asgari2021deep}
\bibinfo{author}{Asgari~Taghanaki, S.}, \bibinfo{author}{Abhishek, K.}, \bibinfo{author}{Cohen, J.~P.}, \bibinfo{author}{Cohen-Adad, J.}, \& \bibinfo{author}{Hamarneh, G.} (\bibinfo{year}{2021}).
\newblock \bibinfo{title}{Deep semantic segmentation of natural and medical images: a review}.
\newblock {\it \bibinfo{journal}{Artificial Intelligence Review}\/},  {\it \bibinfo{volume}{54}\/}, \bibinfo{pages}{137--178}. \DOIprefix\doi{https://doi.org/10.1007/s10462-020-09854-1}.
\bibitem[{Bertasius et~al.(2015)Bertasius, Shi \& Torresani}]{bertasius2015deepedge}
\bibinfo{author}{Bertasius, G.}, \bibinfo{author}{Shi, J.}, \& \bibinfo{author}{Torresani, L.} (\bibinfo{year}{2015}).
\newblock \bibinfo{title}{Deepedge: A multi-scale bifurcated deep network for top-down contour detection}.
\newblock In {\it \bibinfo{booktitle}{Proceedings of the IEEE Conference on Computer Vision and Pattern Recognition}\/} (pp. \bibinfo{pages}{4380--4389}).
\bibitem[{Bhat et~al.(2021)Bhat, Alhashim \& Wonka}]{bhat2021adabins}
\bibinfo{author}{Bhat, S.~F.}, \bibinfo{author}{Alhashim, I.}, \& \bibinfo{author}{Wonka, P.} (\bibinfo{year}{2021}).
\newblock \bibinfo{title}{Adabins: Depth estimation using adaptive bins}.
\newblock In {\it \bibinfo{booktitle}{Proceedings of the IEEE/CVF Conference on Computer Vision and Pattern Recognition}\/} (pp. \bibinfo{pages}{4009--4018}).
\bibitem[{Byeon et~al.(2015)Byeon, Breuel, Raue \& Liwicki}]{byeon2015scene}
\bibinfo{author}{Byeon, W.}, \bibinfo{author}{Breuel, T.~M.}, \bibinfo{author}{Raue, F.}, \& \bibinfo{author}{Liwicki, M.} (\bibinfo{year}{2015}).
\newblock \bibinfo{title}{Scene labeling with lstm recurrent neural networks}.
\newblock In {\it \bibinfo{booktitle}{Proceedings of the IEEE Conference on Computer Vision and Pattern Recognition}\/} (pp. \bibinfo{pages}{3547--3555}).
\bibitem[{Cai et~al.(2022)Cai, Dai, Wang, Chen \& Li}]{cai2021dlnet}
\bibinfo{author}{Cai, Y.}, \bibinfo{author}{Dai, L.}, \bibinfo{author}{Wang, H.}, \bibinfo{author}{Chen, L.}, \& \bibinfo{author}{Li, Y.} (\bibinfo{year}{2022}).
\newblock \bibinfo{title}{Dlnet with training task conversion stream for precise semantic segmentation in actual traffic scene}.
\newblock {\it \bibinfo{journal}{IEEE Transactions on Neural Networks and Learning Systems}\/},  {\it \bibinfo{volume}{33}\/}, \bibinfo{pages}{6443--6457}. \DOIprefix\doi{https://doi.org/10.1109/TNNLS.2021.3080261}.
\bibitem[{Cao et~al.(2023)Cao, Zhang, Lu, Chen, Xiao \& Wang}]{cao2023adaptive}
\bibinfo{author}{Cao, Y.}, \bibinfo{author}{Zhang, H.}, \bibinfo{author}{Lu, X.}, \bibinfo{author}{Chen, Y.}, \bibinfo{author}{Xiao, Z.}, \& \bibinfo{author}{Wang, Y.} (\bibinfo{year}{2023}).
\newblock \bibinfo{title}{Adaptive refining-aggregation-separation framework for unsupervised domain adaptation semantic segmentation}.
\newblock {\it \bibinfo{journal}{IEEE Transactions on Circuits and Systems for Video Technology}\/},  {\it \bibinfo{volume}{33}\/}, \bibinfo{pages}{3822--3832}. \DOIprefix\doi{https://doi.org/10.1109/TCSVT.2023.3243402}.
\bibitem[{Chen et~al.(2018)Chen, Papandreou, Kokkinos, Murphy \& Yuille}]{chen2017deeplab}
\bibinfo{author}{Chen, L.-C.}, \bibinfo{author}{Papandreou, G.}, \bibinfo{author}{Kokkinos, I.}, \bibinfo{author}{Murphy, K.}, \& \bibinfo{author}{Yuille, A.~L.} (\bibinfo{year}{2018}).
\newblock \bibinfo{title}{Deeplab: Semantic image segmentation with deep convolutional nets, atrous convolution, and fully connected crfs}.
\newblock {\it \bibinfo{journal}{IEEE Transactions on Pattern Analysis and Machine Intelligence}\/},  {\it \bibinfo{volume}{40}\/}, \bibinfo{pages}{834--848}. \DOIprefix\doi{https://doi.org/10.1109/TPAMI.2017.2699184}.
\bibitem[{Chen et~al.(2019{\natexlab{a}})Chen, Xue \& Cai}]{chen2019domain}
\bibinfo{author}{Chen, M.}, \bibinfo{author}{Xue, H.}, \& \bibinfo{author}{Cai, D.} (\bibinfo{year}{2019}{\natexlab{a}}).
\newblock \bibinfo{title}{Domain adaptation for semantic segmentation with maximum squares loss}.
\newblock In {\it \bibinfo{booktitle}{Proceedings of the IEEE/CVF International Conference on Computer Vision}\/} (pp. \bibinfo{pages}{2090--2099}).
\bibitem[{Chen et~al.(2022)Chen, Wang, Wang, Zhang, Xie \& Tang}]{chen2021enhanced}
\bibinfo{author}{Chen, T.}, \bibinfo{author}{Wang, S.-H.}, \bibinfo{author}{Wang, Q.}, \bibinfo{author}{Zhang, Z.}, \bibinfo{author}{Xie, G.-S.}, \& \bibinfo{author}{Tang, Z.} (\bibinfo{year}{2022}).
\newblock \bibinfo{title}{Enhanced feature alignment for unsupervised domain adaptation of semantic segmentation}.
\newblock {\it \bibinfo{journal}{IEEE Transactions on Multimedia}\/},  {\it \bibinfo{volume}{24}\/}, \bibinfo{pages}{1042--1054}. \DOIprefix\doi{https://doi.org/10.1109/TMM.2021.3106095}.
\bibitem[{Chen et~al.(2019{\natexlab{b}})Chen, Li, Chen \& Gool}]{chen2019learning}
\bibinfo{author}{Chen, Y.}, \bibinfo{author}{Li, W.}, \bibinfo{author}{Chen, X.}, \& \bibinfo{author}{Gool, L.~V.} (\bibinfo{year}{2019}{\natexlab{b}}).
\newblock \bibinfo{title}{Learning semantic segmentation from synthetic data: A geometrically guided input-output adaptation approach}.
\newblock In {\it \bibinfo{booktitle}{Proceedings of the IEEE/CVF Conference on Computer Vision and Pattern Recognition}\/} (pp. \bibinfo{pages}{1841--1850}).
\bibitem[{Chen et~al.(2017)Chen, Chen, Chen, Tsai, Frank~Wang \& Sun}]{chen2017no}
\bibinfo{author}{Chen, Y.-H.}, \bibinfo{author}{Chen, W.-Y.}, \bibinfo{author}{Chen, Y.-T.}, \bibinfo{author}{Tsai, B.-C.}, \bibinfo{author}{Frank~Wang, Y.-C.}, \& \bibinfo{author}{Sun, M.} (\bibinfo{year}{2017}).
\newblock \bibinfo{title}{No more discrimination: Cross city adaptation of road scene segmenters}.
\newblock In {\it \bibinfo{booktitle}{Proceedings of the IEEE International Conference on Computer Vision}\/} (pp. \bibinfo{pages}{1992--2001}).
\bibitem[{Cordts et~al.(2016)Cordts, Omran, Ramos, Rehfeld, Enzweiler, Benenson, Franke, Roth \& Schiele}]{cordts2016cityscapes}
\bibinfo{author}{Cordts, M.}, \bibinfo{author}{Omran, M.}, \bibinfo{author}{Ramos, S.}, \bibinfo{author}{Rehfeld, T.}, \bibinfo{author}{Enzweiler, M.}, \bibinfo{author}{Benenson, R.}, \bibinfo{author}{Franke, U.}, \bibinfo{author}{Roth, S.}, \& \bibinfo{author}{Schiele, B.} (\bibinfo{year}{2016}).
\newblock \bibinfo{title}{The cityscapes dataset for semantic urban scene understanding}.
\newblock In {\it \bibinfo{booktitle}{Proceedings of the IEEE Conference on Computer Vision and Pattern Recognition}\/} (pp. \bibinfo{pages}{3213--3223}).
\bibitem[{Dong et~al.(2020)Dong, Cong, Sun, Liu \& Xu}]{dong2020cscl}
\bibinfo{author}{Dong, J.}, \bibinfo{author}{Cong, Y.}, \bibinfo{author}{Sun, G.}, \bibinfo{author}{Liu, Y.}, \& \bibinfo{author}{Xu, X.} (\bibinfo{year}{2020}).
\newblock \bibinfo{title}{Cscl: Critical semantic-consistent learning for unsupervised domain adaptation}.
\newblock In {\it \bibinfo{booktitle}{Proceedings of the European Conference on Computer Vision}\/} (pp. \bibinfo{pages}{745--762}).
\newblock \bibinfo{organization}{Springer}.
\bibitem[{Dosovitskiy et~al.(2020)Dosovitskiy, Beyer, Kolesnikov, Weissenborn, Zhai, Unterthiner, Dehghani, Minderer, Heigold, Gelly et~al.}]{dosovitskiy2020image}
\bibinfo{author}{Dosovitskiy, A.}, \bibinfo{author}{Beyer, L.}, \bibinfo{author}{Kolesnikov, A.}, \bibinfo{author}{Weissenborn, D.}, \bibinfo{author}{Zhai, X.}, \bibinfo{author}{Unterthiner, T.}, \bibinfo{author}{Dehghani, M.}, \bibinfo{author}{Minderer, M.}, \bibinfo{author}{Heigold, G.}, \bibinfo{author}{Gelly, S.} et~al. (\bibinfo{year}{2020}).
\newblock \bibinfo{title}{An image is worth 16x16 words: Transformers for image recognition at scale}.
\newblock {\it \bibinfo{journal}{arXiv preprint arXiv:2010.11929}\/}, . \DOIprefix\doi{https://doi.org/10.48550/arXiv.2010.11929}.
\bibitem[{Du et~al.(2019)Du, Tan, Yang, Feng, Xue, Zheng, Ye \& Zhang}]{du2019ssf}
\bibinfo{author}{Du, L.}, \bibinfo{author}{Tan, J.}, \bibinfo{author}{Yang, H.}, \bibinfo{author}{Feng, J.}, \bibinfo{author}{Xue, X.}, \bibinfo{author}{Zheng, Q.}, \bibinfo{author}{Ye, X.}, \& \bibinfo{author}{Zhang, X.} (\bibinfo{year}{2019}).
\newblock \bibinfo{title}{Ssf-dan: Separated semantic feature based domain adaptation network for semantic segmentation}.
\newblock In {\it \bibinfo{booktitle}{Proceedings of the IEEE/CVF International Conference on Computer Vision}\/} (pp. \bibinfo{pages}{982--991}).
\bibitem[{Fu et~al.(2018)Fu, Gong, Wang, Batmanghelich \& Tao}]{fu2018deep}
\bibinfo{author}{Fu, H.}, \bibinfo{author}{Gong, M.}, \bibinfo{author}{Wang, C.}, \bibinfo{author}{Batmanghelich, K.}, \& \bibinfo{author}{Tao, D.} (\bibinfo{year}{2018}).
\newblock \bibinfo{title}{Deep ordinal regression network for monocular depth estimation}.
\newblock In {\it \bibinfo{booktitle}{Proceedings of the IEEE Conference on Computer Vision and Pattern Recognition}\/} (pp. \bibinfo{pages}{2002--2011}).
\bibitem[{Ganin \& Lempitsky(2015)}]{ganin2015unsupervised}
\bibinfo{author}{Ganin, Y.}, \& \bibinfo{author}{Lempitsky, V.} (\bibinfo{year}{2015}).
\newblock \bibinfo{title}{Unsupervised domain adaptation by backpropagation}.
\newblock In {\it \bibinfo{booktitle}{Proceedings of the International Conference on Machine Learning}\/} (pp. \bibinfo{pages}{1180--1189}).
\newblock \bibinfo{organization}{PMLR}.
\bibitem[{Gao et~al.(2022)Gao, Guo, Wang \& Zhang}]{gao2022cross}
\bibinfo{author}{Gao, H.}, \bibinfo{author}{Guo, J.}, \bibinfo{author}{Wang, G.}, \& \bibinfo{author}{Zhang, Q.} (\bibinfo{year}{2022}).
\newblock \bibinfo{title}{Cross-domain correlation distillation for unsupervised domain adaptation in nighttime semantic segmentation}.
\newblock In {\it \bibinfo{booktitle}{Proceedings of the IEEE/CVF Conference on Computer Vision and Pattern Recognition}\/} (pp. \bibinfo{pages}{9913--9923}).
\bibitem[{Gao et~al.(2021)Gao, Zhang \& Zhang}]{gao2021addressing}
\bibinfo{author}{Gao, L.}, \bibinfo{author}{Zhang, L.}, \& \bibinfo{author}{Zhang, Q.} (\bibinfo{year}{2021}).
\newblock \bibinfo{title}{Addressing domain gap via content invariant representation for semantic segmentation}.
\newblock In {\it \bibinfo{booktitle}{Proceedings of the AAAI Conference on Artificial Intelligence}\/} (pp. \bibinfo{pages}{7528--7536}).
\newblock volume~\bibinfo{volume}{35}.
\bibitem[{Goodfellow et~al.(2020)Goodfellow, Pouget-Abadie, Mirza, Xu, Warde-Farley, Ozair, Courville \& Bengio}]{goodfellow2020generative}
\bibinfo{author}{Goodfellow, I.}, \bibinfo{author}{Pouget-Abadie, J.}, \bibinfo{author}{Mirza, M.}, \bibinfo{author}{Xu, B.}, \bibinfo{author}{Warde-Farley, D.}, \bibinfo{author}{Ozair, S.}, \bibinfo{author}{Courville, A.}, \& \bibinfo{author}{Bengio, Y.} (\bibinfo{year}{2020}).
\newblock \bibinfo{title}{Generative adversarial networks}.
\newblock {\it \bibinfo{journal}{Communications of the ACM}\/},  {\it \bibinfo{volume}{63}\/}, \bibinfo{pages}{139--144}. \DOIprefix\doi{https://doi.org/10.1145/3422622}.
\bibitem[{Guo et~al.(2021{\natexlab{a}})Guo, Niu, Qu \& Li}]{guo2021sotr}
\bibinfo{author}{Guo, R.}, \bibinfo{author}{Niu, D.}, \bibinfo{author}{Qu, L.}, \& \bibinfo{author}{Li, Z.} (\bibinfo{year}{2021}{\natexlab{a}}).
\newblock \bibinfo{title}{Sotr: Segmenting objects with transformers}.
\newblock In {\it \bibinfo{booktitle}{Proceedings of the IEEE/CVF International Conference on Computer Vision}\/} (pp. \bibinfo{pages}{7157--7166}).
\bibitem[{Guo et~al.(2021{\natexlab{b}})Guo, Yang, Li \& Yuan}]{guo2021metacorrection}
\bibinfo{author}{Guo, X.}, \bibinfo{author}{Yang, C.}, \bibinfo{author}{Li, B.}, \& \bibinfo{author}{Yuan, Y.} (\bibinfo{year}{2021}{\natexlab{b}}).
\newblock \bibinfo{title}{Metacorrection: Domain-aware meta loss correction for unsupervised domain adaptation in semantic segmentation}.
\newblock In {\it \bibinfo{booktitle}{Proceedings of the IEEE/CVF Conference on Computer Vision and Pattern Recognition}\/} (pp. \bibinfo{pages}{3927--3936}).
\bibitem[{Guo et~al.(2021{\natexlab{c}})Guo, Wang, Hu, Liu, Liu \& Bennamoun}]{guo2020deep}
\bibinfo{author}{Guo, Y.}, \bibinfo{author}{Wang, H.}, \bibinfo{author}{Hu, Q.}, \bibinfo{author}{Liu, H.}, \bibinfo{author}{Liu, L.}, \& \bibinfo{author}{Bennamoun, M.} (\bibinfo{year}{2021}{\natexlab{c}}).
\newblock \bibinfo{title}{Deep learning for 3d point clouds: A survey}.
\newblock {\it \bibinfo{journal}{IEEE Transactions on Pattern Analysis and Machine Intelligence}\/},  {\it \bibinfo{volume}{43}\/}, \bibinfo{pages}{4338--4364}. \DOIprefix\doi{https://doi.org/10.1109/TPAMI.2020.3005434}.
\bibitem[{Hariharan et~al.(2011)Hariharan, Arbel{\'a}ez, Bourdev, Maji \& Malik}]{hariharan2011semantic}
\bibinfo{author}{Hariharan, B.}, \bibinfo{author}{Arbel{\'a}ez, P.}, \bibinfo{author}{Bourdev, L.}, \bibinfo{author}{Maji, S.}, \& \bibinfo{author}{Malik, J.} (\bibinfo{year}{2011}).
\newblock \bibinfo{title}{Semantic contours from inverse detectors}.
\newblock In {\it \bibinfo{booktitle}{2011 International Conference on Computer Vision}\/} (pp. \bibinfo{pages}{991--998}).
\newblock \bibinfo{organization}{IEEE}.
\bibitem[{He et~al.(2016)He, Zhang, Ren \& Sun}]{he2016deep}
\bibinfo{author}{He, K.}, \bibinfo{author}{Zhang, X.}, \bibinfo{author}{Ren, S.}, \& \bibinfo{author}{Sun, J.} (\bibinfo{year}{2016}).
\newblock \bibinfo{title}{Deep residual learning for image recognition}.
\newblock In {\it \bibinfo{booktitle}{Proceedings of the IEEE Conference on Computer Vision and Pattern Recognition}\/} (pp. \bibinfo{pages}{770--778}).
\bibitem[{Hinton et~al.(2015)Hinton, Vinyals, Dean et~al.}]{hinton2015distilling}
\bibinfo{author}{Hinton, G.}, \bibinfo{author}{Vinyals, O.}, \bibinfo{author}{Dean, J.} et~al. (\bibinfo{year}{2015}).
\newblock \bibinfo{title}{Distilling the knowledge in a neural network}.
\newblock {\it \bibinfo{journal}{arXiv preprint arXiv:1503.02531}\/},  {\it \bibinfo{volume}{2}\/}. \DOIprefix\doi{https://doi.org/10.48550/arXiv.1503.02531}.
\bibitem[{Hoffman et~al.(2018)Hoffman, Tzeng, Park, Zhu, Isola, Saenko, Efros \& Darrell}]{hoffman2018cycada}
\bibinfo{author}{Hoffman, J.}, \bibinfo{author}{Tzeng, E.}, \bibinfo{author}{Park, T.}, \bibinfo{author}{Zhu, J.-Y.}, \bibinfo{author}{Isola, P.}, \bibinfo{author}{Saenko, K.}, \bibinfo{author}{Efros, A.}, \& \bibinfo{author}{Darrell, T.} (\bibinfo{year}{2018}).
\newblock \bibinfo{title}{Cycada: Cycle-consistent adversarial domain adaptation}.
\newblock In {\it \bibinfo{booktitle}{Proceedings of the International Conference on Machine Learning (ICML)}\/} (pp. \bibinfo{pages}{1989--1998}).
\newblock \bibinfo{organization}{PMLR}.
\bibitem[{Hoffman et~al.(2016)Hoffman, Wang, Yu \& Darrell}]{hoffman2016fcns}
\bibinfo{author}{Hoffman, J.}, \bibinfo{author}{Wang, D.}, \bibinfo{author}{Yu, F.}, \& \bibinfo{author}{Darrell, T.} (\bibinfo{year}{2016}).
\newblock \bibinfo{title}{Fcns in the wild: Pixel-level adversarial and constraint-based adaptation}.
\newblock {\it \bibinfo{journal}{arXiv preprint arXiv:1612.02649}\/}, . \DOIprefix\doi{https://doi.org/10.48550/arXiv.1612.02649}.
\bibitem[{Hung et~al.(2018)Hung, Tsai, Liou, Lin \& Yang}]{hung2018adversarial}
\bibinfo{author}{Hung, W.-C.}, \bibinfo{author}{Tsai, Y.-H.}, \bibinfo{author}{Liou, Y.-T.}, \bibinfo{author}{Lin, Y.-Y.}, \& \bibinfo{author}{Yang, M.-H.} (\bibinfo{year}{2018}).
\newblock \bibinfo{title}{Adversarial learning for semi-supervised semantic segmentation}.
\newblock {\it \bibinfo{journal}{arXiv preprint arXiv:1802.07934}\/}, . \DOIprefix\doi{https://doi.org/10.48550/arXiv.1802.07934}.
\bibitem[{Jiang(2008)}]{jiang2008literature}
\bibinfo{author}{Jiang, J.} (\bibinfo{year}{2008}).
\newblock \bibinfo{title}{A literature survey on domain adaptation of statistical classifiers}.
\newblock {\it \bibinfo{journal}{URL: http://sifaka.cs.uiuc.edu/jiang4/domainadaptation/survey}\/},  {\it \bibinfo{volume}{3}\/}, \bibinfo{pages}{3}.
\bibitem[{Jiang et~al.(2022)Jiang, Li, Yang, Gao, Wang, Tai \& Wang}]{jiang2022prototypical}
\bibinfo{author}{Jiang, Z.}, \bibinfo{author}{Li, Y.}, \bibinfo{author}{Yang, C.}, \bibinfo{author}{Gao, P.}, \bibinfo{author}{Wang, Y.}, \bibinfo{author}{Tai, Y.}, \& \bibinfo{author}{Wang, C.} (\bibinfo{year}{2022}).
\newblock \bibinfo{title}{Prototypical contrast adaptation for domain adaptive semantic segmentation}.
\newblock In {\it \bibinfo{booktitle}{Proceedings of the European Conference on Computer Vision}\/} (pp. \bibinfo{pages}{36--54}).
\newblock \bibinfo{organization}{Springer}.
\bibitem[{Khoreva et~al.(2016)Khoreva, Benenson, Omran, Hein \& Schiele}]{khoreva2016weakly}
\bibinfo{author}{Khoreva, A.}, \bibinfo{author}{Benenson, R.}, \bibinfo{author}{Omran, M.}, \bibinfo{author}{Hein, M.}, \& \bibinfo{author}{Schiele, B.} (\bibinfo{year}{2016}).
\newblock \bibinfo{title}{Weakly supervised object boundaries}.
\newblock In {\it \bibinfo{booktitle}{Proceedings of the IEEE Conference on Computer Vision and Pattern Recognition}\/} (pp. \bibinfo{pages}{183--192}).
\bibitem[{Klingner et~al.(2022)Klingner, Ayache \& Fingscheidt}]{klingner2022continual}
\bibinfo{author}{Klingner, M.}, \bibinfo{author}{Ayache, M.}, \& \bibinfo{author}{Fingscheidt, T.} (\bibinfo{year}{2022}).
\newblock \bibinfo{title}{Continual batchnorm adaptation (cbna) for semantic segmentation}.
\newblock {\it \bibinfo{journal}{IEEE Transactions on Intelligent Transportation Systems}\/},  {\it \bibinfo{volume}{23}\/}, \bibinfo{pages}{20899--20911}. \DOIprefix\doi{https://doi.org/10.1109/TITS.2022.3190263}.
\bibitem[{Kong et~al.(2022)Kong, Hu, Liu, Lu, You \& Liu}]{kong2022constraining}
\bibinfo{author}{Kong, L.}, \bibinfo{author}{Hu, B.}, \bibinfo{author}{Liu, X.}, \bibinfo{author}{Lu, J.}, \bibinfo{author}{You, J.}, \& \bibinfo{author}{Liu, X.} (\bibinfo{year}{2022}).
\newblock \bibinfo{title}{Constraining pseudo-label in self-training unsupervised domain adaptation with energy-based model}.
\newblock {\it \bibinfo{journal}{International Journal of Intelligent Systems}\/},  {\it \bibinfo{volume}{37}\/}, \bibinfo{pages}{8092--8112}. \DOIprefix\doi{https://doi.org/10.1002/int.22930}.
\bibitem[{Krizhevsky et~al.(2017)Krizhevsky, Sutskever \& Hinton}]{krizhevsky2017imagenet}
\bibinfo{author}{Krizhevsky, A.}, \bibinfo{author}{Sutskever, I.}, \& \bibinfo{author}{Hinton, G.~E.} (\bibinfo{year}{2017}).
\newblock \bibinfo{title}{Imagenet classification with deep convolutional neural networks}.
\newblock {\it \bibinfo{journal}{Communications of the ACM}\/},  {\it \bibinfo{volume}{60}\/}, \bibinfo{pages}{84--90}. \DOIprefix\doi{https://doi.org/10.1145/3065386}.
\bibitem[{Kumar et~al.(2022)Kumar, Kumar \& Lee}]{kumar2022semantic}
\bibinfo{author}{Kumar, S.}, \bibinfo{author}{Kumar, A.}, \& \bibinfo{author}{Lee, D.-G.} (\bibinfo{year}{2022}).
\newblock \bibinfo{title}{Semantic segmentation of uav images based on transformer framework with context information}.
\newblock {\it \bibinfo{journal}{Mathematics}\/},  {\it \bibinfo{volume}{10}\/}, \bibinfo{pages}{4735}. \DOIprefix\doi{https://doi.org/10.3390/math10244735}.
\bibitem[{Lee(2021)}]{lee2021fast}
\bibinfo{author}{Lee, D.-G.} (\bibinfo{year}{2021}).
\newblock \bibinfo{title}{Fast drivable areas estimation with multi-task learning for real-time autonomous driving assistant}.
\newblock {\it \bibinfo{journal}{Applied Sciences}\/},  {\it \bibinfo{volume}{11}\/}, \bibinfo{pages}{10713}. \DOIprefix\doi{https://doi.org/10.3390/app112210713}.
\bibitem[{Lee \& Kim(2022)}]{lee2022joint}
\bibinfo{author}{Lee, D.-G.}, \& \bibinfo{author}{Kim, Y.-K.} (\bibinfo{year}{2022}).
\newblock \bibinfo{title}{Joint semantic understanding with a multilevel branch for driving perception}.
\newblock {\it \bibinfo{journal}{Applied Sciences}\/},  {\it \bibinfo{volume}{12}\/}, \bibinfo{pages}{2877}. \DOIprefix\doi{https://doi.org/10.3390/app12062877}.
\bibitem[{Lee et~al.(2018)Lee, Ros, Li \& Gaidon}]{lee2018spigan}
\bibinfo{author}{Lee, K.-H.}, \bibinfo{author}{Ros, G.}, \bibinfo{author}{Li, J.}, \& \bibinfo{author}{Gaidon, A.} (\bibinfo{year}{2018}).
\newblock \bibinfo{title}{Spigan: Privileged adversarial learning from simulation}.
\newblock {\it \bibinfo{journal}{International Conference on Learning Representations}\/}, .
\bibitem[{Li et~al.(2022)Li, Zhou, Qian, Li, Duan \& Gao}]{li2022feature}
\bibinfo{author}{Li, J.}, \bibinfo{author}{Zhou, K.}, \bibinfo{author}{Qian, S.}, \bibinfo{author}{Li, W.}, \bibinfo{author}{Duan, L.}, \& \bibinfo{author}{Gao, S.} (\bibinfo{year}{2022}).
\newblock \bibinfo{title}{Feature re-representation and reliable pseudo label retraining for cross-domain semantic segmentation}.
\newblock {\it \bibinfo{journal}{IEEE Transactions on Pattern Analysis and Machine Intelligence}\/}, . \DOIprefix\doi{https://doi.org/10.1109/TPAMI.2022.3154933}.
\bibitem[{Lian et~al.(2019)Lian, Lv, Duan \& Gong}]{lian2019constructing}
\bibinfo{author}{Lian, Q.}, \bibinfo{author}{Lv, F.}, \bibinfo{author}{Duan, L.}, \& \bibinfo{author}{Gong, B.} (\bibinfo{year}{2019}).
\newblock \bibinfo{title}{Constructing self-motivated pyramid curriculums for cross-domain semantic segmentation: A non-adversarial approach}.
\newblock In {\it \bibinfo{booktitle}{Proceedings of the IEEE/CVF International Conference on Computer Vision}\/} (pp. \bibinfo{pages}{6758--6767}).
\bibitem[{Liang et~al.(2016)Liang, Shen, Feng, Lin \& Yan}]{liang2016semantic}
\bibinfo{author}{Liang, X.}, \bibinfo{author}{Shen, X.}, \bibinfo{author}{Feng, J.}, \bibinfo{author}{Lin, L.}, \& \bibinfo{author}{Yan, S.} (\bibinfo{year}{2016}).
\newblock \bibinfo{title}{Semantic object parsing with graph lstm}.
\newblock In {\it \bibinfo{booktitle}{Proceedings of the European Conference on Computer Vision}\/} (pp. \bibinfo{pages}{125--143}).
\newblock \bibinfo{organization}{Springer}.
\bibitem[{Long et~al.(2015{\natexlab{a}})Long, Shelhamer \& Darrell}]{long2015fully}
\bibinfo{author}{Long, J.}, \bibinfo{author}{Shelhamer, E.}, \& \bibinfo{author}{Darrell, T.} (\bibinfo{year}{2015}{\natexlab{a}}).
\newblock \bibinfo{title}{Fully convolutional networks for semantic segmentation}.
\newblock In {\it \bibinfo{booktitle}{Proceedings of the IEEE Conference on Computer Vision and Pattern Recognition}\/} (pp. \bibinfo{pages}{3431--3440}).
\bibitem[{Long et~al.(2015{\natexlab{b}})Long, Cao, Wang \& Jordan}]{long2015learning}
\bibinfo{author}{Long, M.}, \bibinfo{author}{Cao, Y.}, \bibinfo{author}{Wang, J.}, \& \bibinfo{author}{Jordan, M.} (\bibinfo{year}{2015}{\natexlab{b}}).
\newblock \bibinfo{title}{Learning transferable features with deep adaptation networks}.
\newblock In {\it \bibinfo{booktitle}{Proceedings of the International Conference on Machine Learning}\/} (pp. \bibinfo{pages}{97--105}).
\newblock \bibinfo{organization}{PMLR}.
\bibitem[{Lv et~al.(2020)Lv, Liang, Chen \& Lin}]{lv2020cross}
\bibinfo{author}{Lv, F.}, \bibinfo{author}{Liang, T.}, \bibinfo{author}{Chen, X.}, \& \bibinfo{author}{Lin, G.} (\bibinfo{year}{2020}).
\newblock \bibinfo{title}{Cross-domain semantic segmentation via domain-invariant interactive relation transfer}.
\newblock In {\it \bibinfo{booktitle}{Proceedings of the IEEE/CVF Conference on Computer Vision and Pattern Recognition}\/} (pp. \bibinfo{pages}{4334--4343}).
\bibitem[{Maninis et~al.(2016)Maninis, Pont-Tuset, Arbel{\'a}ez \& Gool}]{maninis2016deep}
\bibinfo{author}{Maninis, K.-K.}, \bibinfo{author}{Pont-Tuset, J.}, \bibinfo{author}{Arbel{\'a}ez, P.}, \& \bibinfo{author}{Gool, L.~V.} (\bibinfo{year}{2016}).
\newblock \bibinfo{title}{Deep retinal image understanding}.
\newblock In {\it \bibinfo{booktitle}{International Conference on Medical Image Computing and Computer-assisted Intervention}\/} (pp. \bibinfo{pages}{140--148}).
\newblock \bibinfo{organization}{Springer}.
\bibitem[{Minaee et~al.(2022)Minaee, Boykov, Porikli, Plaza, Kehtarnavaz \& Terzopoulos}]{minaee2021image}
\bibinfo{author}{Minaee, S.}, \bibinfo{author}{Boykov, Y.~Y.}, \bibinfo{author}{Porikli, F.}, \bibinfo{author}{Plaza, A.~J.}, \bibinfo{author}{Kehtarnavaz, N.}, \& \bibinfo{author}{Terzopoulos, D.} (\bibinfo{year}{2022}).
\newblock \bibinfo{title}{Image segmentation using deep learning: A survey}.
\newblock {\it \bibinfo{journal}{IEEE Transactions on Pattern Analysis and Machine Intelligence}\/},  {\it \bibinfo{volume}{44}\/}, \bibinfo{pages}{3523--3542}. \DOIprefix\doi{https://doi.org/10.1109/TPAMI.2021.3059968}.
\bibitem[{Murez et~al.(2018)Murez, Kolouri, Kriegman, Ramamoorthi \& Kim}]{murez2018image}
\bibinfo{author}{Murez, Z.}, \bibinfo{author}{Kolouri, S.}, \bibinfo{author}{Kriegman, D.}, \bibinfo{author}{Ramamoorthi, R.}, \& \bibinfo{author}{Kim, K.} (\bibinfo{year}{2018}).
\newblock \bibinfo{title}{Image to image translation for domain adaptation}.
\newblock In {\it \bibinfo{booktitle}{Proceedings of the IEEE Conference on Computer Vision and Pattern Recognition}\/} (pp. \bibinfo{pages}{4500--4509}).
\bibitem[{Musto \& Zinelli(2020)}]{musto2020semantically}
\bibinfo{author}{Musto, L.}, \& \bibinfo{author}{Zinelli, A.} (\bibinfo{year}{2020}).
\newblock \bibinfo{title}{Semantically adaptive image-to-image translation for domain adaptation of semantic segmentation}.
\newblock {\it \bibinfo{journal}{arXiv preprint arXiv:2009.01166}\/}, . \DOIprefix\doi{https://doi.org/10.48550/arXiv.1511.06434}.
\bibitem[{Neuhold et~al.(2017)Neuhold, Ollmann, Rota~Bulo \& Kontschieder}]{neuhold2017mapillary}
\bibinfo{author}{Neuhold, G.}, \bibinfo{author}{Ollmann, T.}, \bibinfo{author}{Rota~Bulo, S.}, \& \bibinfo{author}{Kontschieder, P.} (\bibinfo{year}{2017}).
\newblock \bibinfo{title}{The mapillary vistas dataset for semantic understanding of street scenes}.
\newblock In {\it \bibinfo{booktitle}{Proceedings of the IEEE International Conference on Computer Vision (ICCV)}\/} (pp. \bibinfo{pages}{4990--4999}).
\bibitem[{Pan et~al.(2020)Pan, Shin, Rameau, Lee \& Kweon}]{pan2020unsupervised}
\bibinfo{author}{Pan, F.}, \bibinfo{author}{Shin, I.}, \bibinfo{author}{Rameau, F.}, \bibinfo{author}{Lee, S.}, \& \bibinfo{author}{Kweon, I.~S.} (\bibinfo{year}{2020}).
\newblock \bibinfo{title}{Unsupervised intra-domain adaptation for semantic segmentation through self-supervision}.
\newblock In {\it \bibinfo{booktitle}{Proceedings of the IEEE/CVF Conference on Computer Vision and Pattern Recognition}\/} (pp. \bibinfo{pages}{3764--3773}).
\bibitem[{Park et~al.(2019)Park, Woo, Kim, Cho \& Kweon}]{park2019preserving}
\bibinfo{author}{Park, K.}, \bibinfo{author}{Woo, S.}, \bibinfo{author}{Kim, D.}, \bibinfo{author}{Cho, D.}, \& \bibinfo{author}{Kweon, I.~S.} (\bibinfo{year}{2019}).
\newblock \bibinfo{title}{Preserving semantic and temporal consistency for unpaired video-to-video translation}.
\newblock In {\it \bibinfo{booktitle}{Proceedings of the 27th ACM International Conference on Multimedia}\/} (pp. \bibinfo{pages}{1248--1257}).
\bibitem[{Poma et~al.(2020)Poma, Riba \& Sappa}]{poma2020dense}
\bibinfo{author}{Poma, X.~S.}, \bibinfo{author}{Riba, E.}, \& \bibinfo{author}{Sappa, A.} (\bibinfo{year}{2020}).
\newblock \bibinfo{title}{Dense extreme inception network: Towards a robust cnn model for edge detection}.
\newblock In {\it \bibinfo{booktitle}{Proceedings of the IEEE/CVF Winter Conference on Applications of Computer Vision}\/} (pp. \bibinfo{pages}{1923--1932}).
\bibitem[{Radford et~al.(2015)Radford, Metz \& Chintala}]{radford2015unsupervised}
\bibinfo{author}{Radford, A.}, \bibinfo{author}{Metz, L.}, \& \bibinfo{author}{Chintala, S.} (\bibinfo{year}{2015}).
\newblock \bibinfo{title}{Unsupervised representation learning with deep convolutional generative adversarial networks}.
\newblock {\it \bibinfo{journal}{arXiv preprint arXiv:1511.06434}\/}, . \DOIprefix\doi{https://doi.org/10.48550/arXiv.1511.06434}.
\bibitem[{Ranftl et~al.(2021)Ranftl, Bochkovskiy \& Koltun}]{ranftl2021vision}
\bibinfo{author}{Ranftl, R.}, \bibinfo{author}{Bochkovskiy, A.}, \& \bibinfo{author}{Koltun, V.} (\bibinfo{year}{2021}).
\newblock \bibinfo{title}{Vision transformers for dense prediction}.
\newblock In {\it \bibinfo{booktitle}{Proceedings of the IEEE/CVF International Conference on Computer Vision}\/} (pp. \bibinfo{pages}{12179--12188}).
\bibitem[{Richter et~al.(2016)Richter, Vineet, Roth \& Koltun}]{richter2016playing}
\bibinfo{author}{Richter, S.~R.}, \bibinfo{author}{Vineet, V.}, \bibinfo{author}{Roth, S.}, \& \bibinfo{author}{Koltun, V.} (\bibinfo{year}{2016}).
\newblock \bibinfo{title}{Playing for data: Ground truth from computer games}.
\newblock In {\it \bibinfo{booktitle}{Proceedings of the European Conference on Computer Vision}\/} (pp. \bibinfo{pages}{102--118}).
\newblock \bibinfo{organization}{Springer}.
\bibitem[{Ronneberger et~al.(2015)Ronneberger, Fischer \& Brox}]{ronneberger2015u}
\bibinfo{author}{Ronneberger, O.}, \bibinfo{author}{Fischer, P.}, \& \bibinfo{author}{Brox, T.} (\bibinfo{year}{2015}).
\newblock \bibinfo{title}{U-net: Convolutional networks for biomedical image segmentation}.
\newblock In {\it \bibinfo{booktitle}{International Conference on Medical Image Computing and Computer-assisted Intervention}\/} (pp. \bibinfo{pages}{234--241}).
\newblock \bibinfo{organization}{Springer}.
\bibitem[{Ros et~al.(2016)Ros, Sellart, Materzynska, Vazquez \& Lopez}]{ros2016synthia}
\bibinfo{author}{Ros, G.}, \bibinfo{author}{Sellart, L.}, \bibinfo{author}{Materzynska, J.}, \bibinfo{author}{Vazquez, D.}, \& \bibinfo{author}{Lopez, A.~M.} (\bibinfo{year}{2016}).
\newblock \bibinfo{title}{The synthia dataset: A large collection of synthetic images for semantic segmentation of urban scenes}.
\newblock In {\it \bibinfo{booktitle}{Proceedings of the IEEE Conference on Computer Vision and Pattern Recognition}\/} (pp. \bibinfo{pages}{3234--3243}).
\bibitem[{Saha et~al.(2021)Saha, Obukhov, Paudel, Kanakis, Chen, Georgoulis \& Van~Gool}]{saha2021learning}
\bibinfo{author}{Saha, S.}, \bibinfo{author}{Obukhov, A.}, \bibinfo{author}{Paudel, D.~P.}, \bibinfo{author}{Kanakis, M.}, \bibinfo{author}{Chen, Y.}, \bibinfo{author}{Georgoulis, S.}, \& \bibinfo{author}{Van~Gool, L.} (\bibinfo{year}{2021}).
\newblock \bibinfo{title}{Learning to relate depth and semantics for unsupervised domain adaptation}.
\newblock In {\it \bibinfo{booktitle}{Proceedings of the IEEE/CVF Conference on Computer Vision and Pattern Recognition}\/} (pp. \bibinfo{pages}{8197--8207}).
\bibitem[{Saito et~al.(2018)Saito, Watanabe, Ushiku \& Harada}]{saito2018maximum}
\bibinfo{author}{Saito, K.}, \bibinfo{author}{Watanabe, K.}, \bibinfo{author}{Ushiku, Y.}, \& \bibinfo{author}{Harada, T.} (\bibinfo{year}{2018}).
\newblock \bibinfo{title}{Maximum classifier discrepancy for unsupervised domain adaptation}.
\newblock In {\it \bibinfo{booktitle}{Proceedings of the IEEE Conference on Computer Vision and Pattern Recognition}\/} (pp. \bibinfo{pages}{3723--3732}).
\bibitem[{Saporta et~al.(2020)Saporta, Vu, Cord \& P{\'e}rez}]{saporta2020esl}
\bibinfo{author}{Saporta, A.}, \bibinfo{author}{Vu, T.-H.}, \bibinfo{author}{Cord, M.}, \& \bibinfo{author}{P{\'e}rez, P.} (\bibinfo{year}{2020}).
\newblock \bibinfo{title}{Esl: Entropy-guided self-supervised learning for domain adaptation in semantic segmentation}.
\newblock {\it \bibinfo{journal}{arXiv preprint arXiv:2006.08658}\/}, . \DOIprefix\doi{https://doi.org/10.48550/arXiv.2006.08658}.
\bibitem[{Simonyan \& Zisserman(2014)}]{simonyan2014very}
\bibinfo{author}{Simonyan, K.}, \& \bibinfo{author}{Zisserman, A.} (\bibinfo{year}{2014}).
\newblock \bibinfo{title}{Very deep convolutional networks for large-scale image recognition}.
\newblock {\it \bibinfo{journal}{arXiv preprint arXiv:1409.1556}\/}, . \DOIprefix\doi{https://doi.org/10.48550/arXiv.1409.1556}.
\bibitem[{Singh \& Singh(2008)}]{singh2008edge}
\bibinfo{author}{Singh, B.}, \& \bibinfo{author}{Singh, A.~P.} (\bibinfo{year}{2008}).
\newblock \bibinfo{title}{Edge detection in gray level images based on the shannon entropy}.
\newblock {\it \bibinfo{journal}{Journal of Computer Science}\/},  {\it \bibinfo{volume}{4}\/}, \bibinfo{pages}{186--191}. \DOIprefix\doi{https://doi.org/10.3844/jcssp.2008.186.191}.
\bibitem[{Stan \& Rostami(2021)}]{stan2021unsupervised}
\bibinfo{author}{Stan, S.}, \& \bibinfo{author}{Rostami, M.} (\bibinfo{year}{2021}).
\newblock \bibinfo{title}{Unsupervised model adaptation for continual semantic segmentation}.
\newblock In {\it \bibinfo{booktitle}{Proceedings of the AAAI Conference on Artificial Intelligence}\/} (pp. \bibinfo{pages}{2593--2601}).
\newblock volume~\bibinfo{volume}{35}.
\bibitem[{Strudel et~al.(2021)Strudel, Garcia, Laptev \& Schmid}]{strudel2021segmenter}
\bibinfo{author}{Strudel, R.}, \bibinfo{author}{Garcia, R.}, \bibinfo{author}{Laptev, I.}, \& \bibinfo{author}{Schmid, C.} (\bibinfo{year}{2021}).
\newblock \bibinfo{title}{Segmenter: Transformer for semantic segmentation}.
\newblock In {\it \bibinfo{booktitle}{Proceedings of the IEEE/CVF International Conference on Computer Vision}\/} (pp. \bibinfo{pages}{7262--7272}).
\bibitem[{Subhani \& Ali(2020)}]{subhani2020learning}
\bibinfo{author}{Subhani, M.~N.}, \& \bibinfo{author}{Ali, M.} (\bibinfo{year}{2020}).
\newblock \bibinfo{title}{Learning from scale-invariant examples for domain adaptation in semantic segmentation}.
\newblock In {\it \bibinfo{booktitle}{Proceedings of the European Conference on Computer Vision}\/} (pp. \bibinfo{pages}{290--306}).
\newblock \bibinfo{organization}{Springer}.
\bibitem[{Takikawa et~al.(2019)Takikawa, Acuna, Jampani \& Fidler}]{takikawa2019gated}
\bibinfo{author}{Takikawa, T.}, \bibinfo{author}{Acuna, D.}, \bibinfo{author}{Jampani, V.}, \& \bibinfo{author}{Fidler, S.} (\bibinfo{year}{2019}).
\newblock \bibinfo{title}{Gated-scnn: Gated shape cnns for semantic segmentation}.
\newblock In {\it \bibinfo{booktitle}{Proceedings of the IEEE/CVF International Conference on Computer Vision}\/} (pp. \bibinfo{pages}{5229--5238}).
\bibitem[{Toldo et~al.(2021)Toldo, Michieli \& Zanuttigh}]{toldo2021unsupervised}
\bibinfo{author}{Toldo, M.}, \bibinfo{author}{Michieli, U.}, \& \bibinfo{author}{Zanuttigh, P.} (\bibinfo{year}{2021}).
\newblock \bibinfo{title}{Unsupervised domain adaptation in semantic segmentation via orthogonal and clustered embeddings}.
\newblock In {\it \bibinfo{booktitle}{Proceedings of the IEEE/CVF Winter Conference on Applications of Computer Vision}\/} (pp. \bibinfo{pages}{1358--1368}).
\bibitem[{Tsai et~al.(2018)Tsai, Hung, Schulter, Sohn, Yang \& Chandraker}]{tsai2018learning}
\bibinfo{author}{Tsai, Y.-H.}, \bibinfo{author}{Hung, W.-C.}, \bibinfo{author}{Schulter, S.}, \bibinfo{author}{Sohn, K.}, \bibinfo{author}{Yang, M.-H.}, \& \bibinfo{author}{Chandraker, M.} (\bibinfo{year}{2018}).
\newblock \bibinfo{title}{Learning to adapt structured output space for semantic segmentation}.
\newblock In {\it \bibinfo{booktitle}{Proceedings of the IEEE Conference on Computer Vision and Pattern Recognition}\/} (pp. \bibinfo{pages}{7472--7481}).
\bibitem[{Tsai et~al.(2019)Tsai, Sohn, Schulter \& Chandraker}]{tsai2019domain}
\bibinfo{author}{Tsai, Y.-H.}, \bibinfo{author}{Sohn, K.}, \bibinfo{author}{Schulter, S.}, \& \bibinfo{author}{Chandraker, M.} (\bibinfo{year}{2019}).
\newblock \bibinfo{title}{Domain adaptation for structured output via discriminative patch representations}.
\newblock In {\it \bibinfo{booktitle}{Proceedings of the IEEE/CVF International Conference on Computer Vision (ICCV)}\/} (pp. \bibinfo{pages}{1456--1465}).
\bibitem[{Visin et~al.(2016)Visin, Ciccone, Romero, Kastner, Cho, Bengio, Matteucci \& Courville}]{visin2016reseg}
\bibinfo{author}{Visin, F.}, \bibinfo{author}{Ciccone, M.}, \bibinfo{author}{Romero, A.}, \bibinfo{author}{Kastner, K.}, \bibinfo{author}{Cho, K.}, \bibinfo{author}{Bengio, Y.}, \bibinfo{author}{Matteucci, M.}, \& \bibinfo{author}{Courville, A.} (\bibinfo{year}{2016}).
\newblock \bibinfo{title}{Reseg: A recurrent neural network-based model for semantic segmentation}.
\newblock In {\it \bibinfo{booktitle}{Proceedings of the IEEE Conference on Computer Vision and Pattern Recognition Workshops}\/} (pp. \bibinfo{pages}{41--48}).
\bibitem[{Vu et~al.(2019{\natexlab{a}})Vu, Jain, Bucher, Cord \& P{\'e}rez}]{vu2019advent}
\bibinfo{author}{Vu, T.-H.}, \bibinfo{author}{Jain, H.}, \bibinfo{author}{Bucher, M.}, \bibinfo{author}{Cord, M.}, \& \bibinfo{author}{P{\'e}rez, P.} (\bibinfo{year}{2019}{\natexlab{a}}).
\newblock \bibinfo{title}{Advent: Adversarial entropy minimization for domain adaptation in semantic segmentation}.
\newblock In {\it \bibinfo{booktitle}{Proceedings of the IEEE/CVF Conference on Computer Vision and Pattern Recognition}\/} (pp. \bibinfo{pages}{2517--2526}).
\bibitem[{Vu et~al.(2019{\natexlab{b}})Vu, Jain, Bucher, Cord \& P{\'e}rez}]{vu2019dada}
\bibinfo{author}{Vu, T.-H.}, \bibinfo{author}{Jain, H.}, \bibinfo{author}{Bucher, M.}, \bibinfo{author}{Cord, M.}, \& \bibinfo{author}{P{\'e}rez, P.} (\bibinfo{year}{2019}{\natexlab{b}}).
\newblock \bibinfo{title}{Dada: Depth-aware domain adaptation in semantic segmentation}.
\newblock In {\it \bibinfo{booktitle}{Proceedings of the IEEE/CVF International Conference on Computer Vision}\/} (pp. \bibinfo{pages}{7364--7373}).
\bibitem[{Wang et~al.(2020)Wang, Shen, Zhang, Duan \& Mei}]{wang2020classes}
\bibinfo{author}{Wang, H.}, \bibinfo{author}{Shen, T.}, \bibinfo{author}{Zhang, W.}, \bibinfo{author}{Duan, L.-Y.}, \& \bibinfo{author}{Mei, T.} (\bibinfo{year}{2020}).
\newblock \bibinfo{title}{Classes matter: A fine-grained adversarial approach to cross-domain semantic segmentation}.
\newblock In {\it \bibinfo{booktitle}{Proceedings of the European Conference on Computer Vision}\/} (pp. \bibinfo{pages}{642--659}).
\newblock \bibinfo{organization}{Springer}.
\bibitem[{Xie et~al.(2021)Xie, Wang, Yu, Anandkumar, Alvarez \& Luo}]{xie2021segformer}
\bibinfo{author}{Xie, E.}, \bibinfo{author}{Wang, W.}, \bibinfo{author}{Yu, Z.}, \bibinfo{author}{Anandkumar, A.}, \bibinfo{author}{Alvarez, J.~M.}, \& \bibinfo{author}{Luo, P.} (\bibinfo{year}{2021}).
\newblock \bibinfo{title}{Segformer: Simple and efficient design for semantic segmentation with transformers}.
\newblock In {\it \bibinfo{booktitle}{Advances in Neural Information Processing Systems}\/} (pp. \bibinfo{pages}{12077--12090}).
\newblock \bibinfo{publisher}{Curran Associates, Inc.} volume~\bibinfo{volume}{34}.
\bibitem[{Yu \& Koltun(2015)}]{yu2015multi}
\bibinfo{author}{Yu, F.}, \& \bibinfo{author}{Koltun, V.} (\bibinfo{year}{2015}).
\newblock \bibinfo{title}{Multi-scale context aggregation by dilated convolutions}.
\newblock {\it \bibinfo{journal}{arXiv preprint arXiv:1511.07122}\/}, . \DOIprefix\doi{https://doi.org/10.48550/arXiv.1511.0712}.
\bibitem[{Yu et~al.(2021)Yu, Zhang, Dong, Hu, Dong \& Zhang}]{yu2021dast}
\bibinfo{author}{Yu, F.}, \bibinfo{author}{Zhang, M.}, \bibinfo{author}{Dong, H.}, \bibinfo{author}{Hu, S.}, \bibinfo{author}{Dong, B.}, \& \bibinfo{author}{Zhang, L.} (\bibinfo{year}{2021}).
\newblock \bibinfo{title}{Dast: Unsupervised domain adaptation in semantic segmentation based on discriminator attention and self-training}.
\newblock In {\it \bibinfo{booktitle}{Proceedings of the AAAI Conference on Artificial Intelligence}\/} (pp. \bibinfo{pages}{10754--10762}).
\newblock volume~\bibinfo{volume}{35}.
\bibitem[{Yu et~al.(2017)Yu, Feng, Liu \& Ramalingam}]{yu2017casenet}
\bibinfo{author}{Yu, Z.}, \bibinfo{author}{Feng, C.}, \bibinfo{author}{Liu, M.-Y.}, \& \bibinfo{author}{Ramalingam, S.} (\bibinfo{year}{2017}).
\newblock \bibinfo{title}{Casenet: Deep category-aware semantic edge detection}.
\newblock In {\it \bibinfo{booktitle}{Proceedings of the IEEE Conference on Computer Vision and Pattern Recognition}\/} (pp. \bibinfo{pages}{5964--5973}).
\bibitem[{Yu et~al.(2018)Yu, Liu, Zou, Feng, Ramalingam, Kumar \& Kautz}]{yu2018simultaneous}
\bibinfo{author}{Yu, Z.}, \bibinfo{author}{Liu, W.}, \bibinfo{author}{Zou, Y.}, \bibinfo{author}{Feng, C.}, \bibinfo{author}{Ramalingam, S.}, \bibinfo{author}{Kumar, B.}, \& \bibinfo{author}{Kautz, J.} (\bibinfo{year}{2018}).
\newblock \bibinfo{title}{Simultaneous edge alignment and learning}.
\newblock In {\it \bibinfo{booktitle}{Proceedings of the European Conference on Computer Vision}\/} (pp. \bibinfo{pages}{388--404}).
\bibitem[{Zhang et~al.(2022)Zhang, Chen, Shen, Shen, Zhang \& Zhang}]{zhang2022confidence}
\bibinfo{author}{Zhang, X.}, \bibinfo{author}{Chen, Y.}, \bibinfo{author}{Shen, Z.}, \bibinfo{author}{Shen, Y.}, \bibinfo{author}{Zhang, H.}, \& \bibinfo{author}{Zhang, Y.} (\bibinfo{year}{2022}).
\newblock \bibinfo{title}{Confidence-and-refinement adaptation model for cross-domain semantic segmentation}.
\newblock {\it \bibinfo{journal}{IEEE Transactions on Intelligent Transportation Systems}\/},  {\it \bibinfo{volume}{23}\/}, \bibinfo{pages}{9529--9542}. \DOIprefix\doi{https://doi.org/10.1109/TITS.2022.3140481}.
\bibitem[{Zhang(2021)}]{zhang2021survey}
\bibinfo{author}{Zhang, Y.} (\bibinfo{year}{2021}).
\newblock \bibinfo{title}{A survey of unsupervised domain adaptation for visual recognition}.
\newblock {\it \bibinfo{journal}{arXiv preprint arXiv:2112.06745}\/}, . \DOIprefix\doi{https://doi.org/10.48550/arXiv.2112.06745}.
\bibitem[{Zhang et~al.(2020)Zhang, David, Foroosh \& Gong}]{zhang2019curriculum}
\bibinfo{author}{Zhang, Y.}, \bibinfo{author}{David, P.}, \bibinfo{author}{Foroosh, H.}, \& \bibinfo{author}{Gong, B.} (\bibinfo{year}{2020}).
\newblock \bibinfo{title}{A curriculum domain adaptation approach to the semantic segmentation of urban scenes}.
\newblock {\it \bibinfo{journal}{IEEE Transactions on Pattern Analysis and Machine Intelligence}\/},  {\it \bibinfo{volume}{42}\/}, \bibinfo{pages}{1823--1841}. \DOIprefix\doi{https://doi.org/10.1109/TPAMI.2019.2903401}.
\bibitem[{Zhang et~al.(2017)Zhang, David \& Gong}]{zhang2017curriculum}
\bibinfo{author}{Zhang, Y.}, \bibinfo{author}{David, P.}, \& \bibinfo{author}{Gong, B.} (\bibinfo{year}{2017}).
\newblock \bibinfo{title}{Curriculum domain adaptation for semantic segmentation of urban scenes}.
\newblock In {\it \bibinfo{booktitle}{Proceedings of the IEEE International Conference on Computer Vision}\/} (pp. \bibinfo{pages}{2020--2030}).
\bibitem[{Zhao et~al.(2017)Zhao, Shi, Qi, Wang \& Jia}]{zhao2017pyramid}
\bibinfo{author}{Zhao, H.}, \bibinfo{author}{Shi, J.}, \bibinfo{author}{Qi, X.}, \bibinfo{author}{Wang, X.}, \& \bibinfo{author}{Jia, J.} (\bibinfo{year}{2017}).
\newblock \bibinfo{title}{Pyramid scene parsing network}.
\newblock In {\it \bibinfo{booktitle}{Proceedings of the IEEE Conference on Computer Vision and Pattern Recognition}\/} (pp. \bibinfo{pages}{2881--2890}).
\bibitem[{Zhao et~al.(2019)Zhao, Zheng, Xu \& Wu}]{zhao2019object}
\bibinfo{author}{Zhao, Z.-Q.}, \bibinfo{author}{Zheng, P.}, \bibinfo{author}{Xu, S.-t.}, \& \bibinfo{author}{Wu, X.} (\bibinfo{year}{2019}).
\newblock \bibinfo{title}{Object detection with deep learning: A review}.
\newblock {\it \bibinfo{journal}{IEEE Transactions on Neural Networks and Learning Systems}\/},  {\it \bibinfo{volume}{30}\/}, \bibinfo{pages}{3212--3232}. \DOIprefix\doi{https://doi.org/10.1109/TNNLS.2018.2876865}.
\bibitem[{Zhou et~al.(2021)Zhou, Wang, Chu, Yang, Bai \& Xu}]{zhou2020affinity}
\bibinfo{author}{Zhou, W.}, \bibinfo{author}{Wang, Y.}, \bibinfo{author}{Chu, J.}, \bibinfo{author}{Yang, J.}, \bibinfo{author}{Bai, X.}, \& \bibinfo{author}{Xu, Y.} (\bibinfo{year}{2021}).
\newblock \bibinfo{title}{Affinity space adaptation for semantic segmentation across domains}.
\newblock {\it \bibinfo{journal}{IEEE Transactions on Image Processing}\/},  {\it \bibinfo{volume}{30}\/}, \bibinfo{pages}{2549--2561}. \DOIprefix\doi{https://doi.org/10.1109/TIP.2020.3018221}.
\bibitem[{Zhu et~al.(2017)Zhu, Park, Isola \& Efros}]{zhu2017unpaired}
\bibinfo{author}{Zhu, J.-Y.}, \bibinfo{author}{Park, T.}, \bibinfo{author}{Isola, P.}, \& \bibinfo{author}{Efros, A.~A.} (\bibinfo{year}{2017}).
\newblock \bibinfo{title}{Unpaired image-to-image translation using cycle-consistent adversarial networks}.
\newblock In {\it \bibinfo{booktitle}{Proceedings of the IEEE International Conference on Computer Vision}\/} (pp. \bibinfo{pages}{2223--2232}).
\bibitem[{Zou et~al.(2018)Zou, Yu, Kumar \& Wang}]{zou2018unsupervised}
\bibinfo{author}{Zou, Y.}, \bibinfo{author}{Yu, Z.}, \bibinfo{author}{Kumar, B.}, \& \bibinfo{author}{Wang, J.} (\bibinfo{year}{2018}).
\newblock \bibinfo{title}{Unsupervised domain adaptation for semantic segmentation via class-balanced self-training}.
\newblock In {\it \bibinfo{booktitle}{Proceedings of the European Conference on Computer Vision}\/} (pp. \bibinfo{pages}{289--305}).

\end{thebibliography}

\end{document}